\documentclass[sigconf]{acmart}

\usepackage{algorithm}
\usepackage{algpseudocode}
\usepackage{multirow}
\usepackage[export]{adjustbox}

\newcommand{\argmin}{\mathop{\mathrm{argmin}}} 
 
\newcommand{\ul}{\underline} 
\newcommand{\bd}{\textbf} 

\AtBeginDocument{%
  \providecommand\BibTeX{{%
    \normalfont B\kern-0.5em{\scshape i\kern-0.25em b}\kern-0.8em\TeX}}}

\setcopyright{acmlicensed}
\copyrightyear{~~~~~~~~~}
\acmYear{~~~~~~}
\acmDOI{XXXXXXX.XXXXXXX}

\acmConference[Preprint]{}{}{}
\acmISBN{~~~~~~~~~~~}

\begin{document}

\title{NFCL: Simply interpretable neural networks for a short-term multivariate forecasting}

\author{Wonkeun Jo}
\email{jowonkun@g.cnu.ac.kr}
\orcid{0000-0001-7374-2900}
\affiliation{%
  \institution{Department of Computer Science and Engineering, Chungnam National University}
  \streetaddress{99 Daehak--ro}
  \city{Yuseong--gu}
  \state{Daejeon}
  \country{Republic of Korea}
  \postcode{34134}
}

\author{Dongil Kim}
\authornote{Corresponding author.}
\email{d.kim@ewha.ac.kr}
\orcid{0000-0001-7425-7579}
\affiliation{%
  \institution{Department of Data Science, 
  \\ Ewha Womans University}
  \streetaddress{52 Ewhayeodae--gil}
  \city{Seodaemun--gu}
  \state{Seoul}
  \country{Republic of Korea}}

\renewcommand{\shortauthors}{Jo and Kim.}

\begin{abstract}
    Multivariate time-series forecasting (MTSF) stands as a compelling field within the machine learning community. 
    Diverse neural network based methodologies deployed in MTSF applications have demonstrated commendable efficacy. 
    Despite the advancements in model performance, comprehending the rationale behind the model’s behavior remains an enigma. 
    Our proposed model, the Neural ForeCasting Layer (NFCL), employs a straightforward amalgamation of neural networks. 
    This uncomplicated integration ensures that each neural network contributes inputs and predictions independently, devoid of interference from other inputs. 
    Consequently, our model facilitates a transparent explication of forecast results. 
    This paper introduces NFCL along with its diverse extensions. 
    Empirical findings underscore NFCL’s superior performance compared to nine benchmark models across 15 available open datasets. 
    Notably, NFCL not only surpasses competitors but also provides elucidation for its predictions. 
    In addition, Rigorous experimentation involving diverse model structures bolsters the justification of NFCL’s unique configuration.
\end{abstract}

\ccsdesc[500]{Computing methodologies~Artificial intelligence}
\ccsdesc[300]{Computing methodologies~Modeling methodologies}



\keywords{Deep learning; Multivariate time-series; Forecasting; Interpretable;}



\maketitle


\balance

\section{Introduction}

Multivariate Time-Series Forecasting (MTSF) plays a crucial role in the machine learning society, significantly impacting various aspects of our lives. 
For instance, machine learning algorithms control the quantity of electricity generated \cite{elec_gen_contol}. 
Additionally, atmospheric conditions are forecasted using neural network (NN) based on past time-series data \cite{weather_forecasting, PM_forecasting, eg_forecasting_wind_01, baqdataset}. 
Given the multitude of applications beyond these examples, the MTSF problem attracts numerous researchers striving for more precise solutions \cite{time-series_survey}.

\begin{figure}[h]
  \centering
  \includegraphics[width=\columnwidth, right]{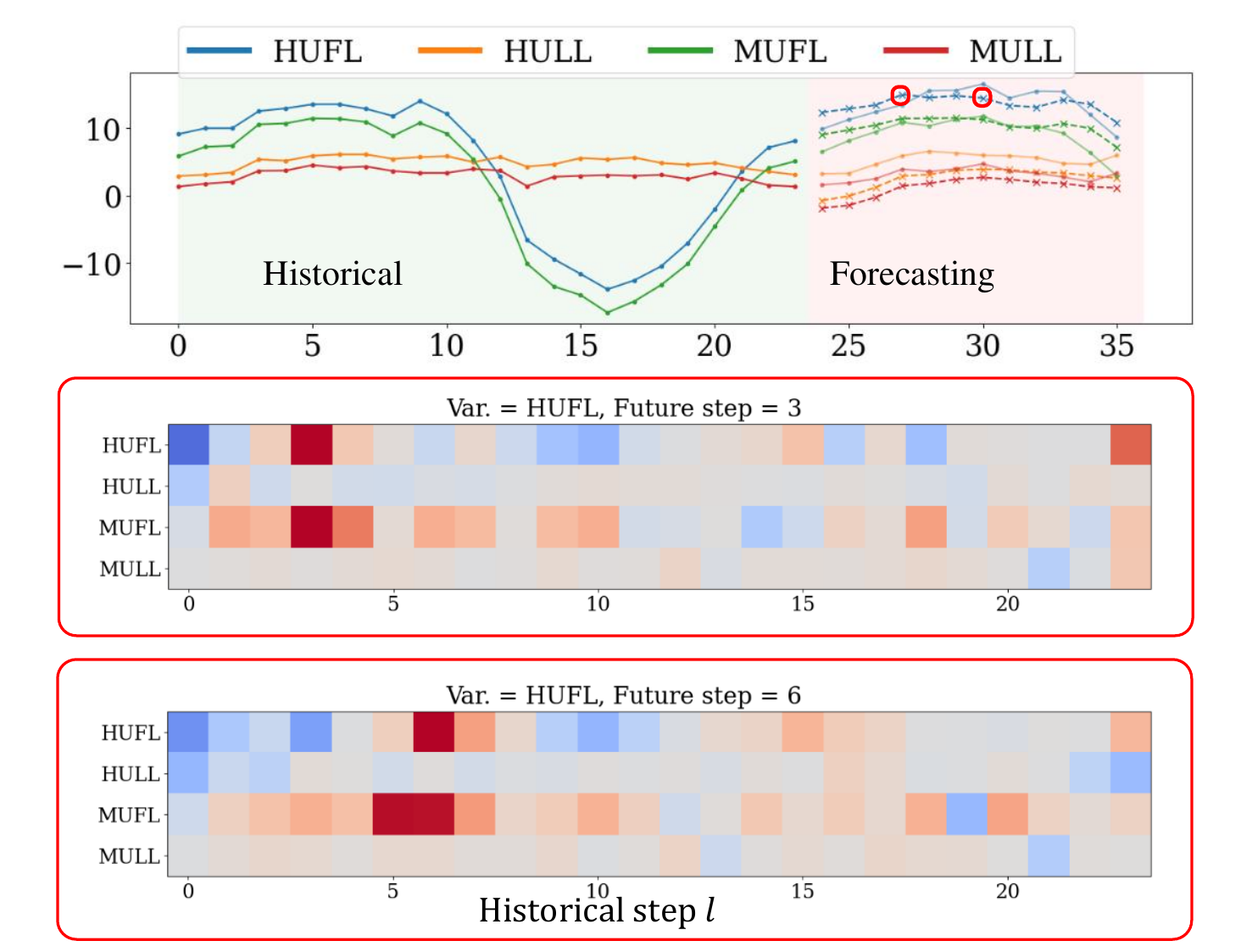}
  \caption{Forecasting the fourth and seventh next steps' HUFL based on the historical multivariate time-series. Below, contribution maps illustrate the reasons why the model predicts the target to be closer to $\hat{Y}^{\mathrm{HUFL}}=10$. The pattern of the contribution map predicted by the model changes, similar to moving three steps forward in the fourth future step while the target time point is moved to three step forward.}
  \label{fig. flow}
  \Description{  }
\end{figure}

NNs are the preferred method for addressing encompassing classification, regression, anomaly detection, and more \cite{ai_app_01, ai_app_02}. 
And, recent approaches 
aim to solve time-series problems, including forecasting \cite{informer, timesnet, dnlinear}. 
However, our focus aligns with two trends in the machine learning society: 1) contemporary research emphasizes performance and architectural complexity \cite{effnet_v2, largest_vit}, and 2) there is a growing need for short-term future forecasting using the latest NNs \cite{eg_forecasting_wind_01, eg_forecasting_elec_01, eg_forecasting_elec_02, eg_forecasting_passen_01}.

Interpretability is another critical aspect in the machine learning community, addressing the need to understand why a method made specific predictions \cite{xai_for_timeseries}. 
In real-world decision-making, users seek information on how black-box ML models generate results, and providing this insight can aid individuals in making informed decisions \cite{exai_time_01, exai_time_02, eg_forecasting_wind_01}. 
A simple method, such as an additive model applying independent functions to each input, represents a solution in machine learning to interpret the reasons for forecasting time-series \cite{dnlinear}.

We propose \textit{\textbf{N}eural \textbf{F}ore\textbf{C}asting \textbf{L}ayer} (NFCL) for forecasting the near future while elucidating the reasons behind predictions. 
Figure \ref{fig. flow} illustrates the concept of NFCL. 
The model receives the input series, depicted in green, and subsequently forecasts the future, represented in red, using contribution maps. 
In these maps, the red color signifies positive weights, while the blue color denotes negative weights.
It is important to note that the grey color, closer to zero, is negligible in the forecasting process.
A simple example demonstrates that when forecasting the High UseFul Load (HUFL) in the next fourth and seventh steps future from the present (because the future time step starts from zero index), NFCL assigns higher positive weights from Middle UseFul Load (MUFL) and mixed weights from the HUFL.
According to the research, Note that the HUFL must be higher than MUFL, and they are correlated \cite{informer}.

We also tested expansion of NFCL, which treats each variable (representing one step of the time-series) independently, by applying various preprocessing methods, such as statistical normalization, input re-evaluation, time-series decomposition and etc.
We illustrate that NFCL achieves the best scores in real-world datasets and elucidate the reasons behind the model's forecasts. Our main contributions are as follows:

\begin{itemize}
    \item NFCL is constructed based on simple weight combinations, such as linear regression. We observe that capturing cross-correlation between NFCL and other time-series contributes more to forecasting than handling only its own time-series independently.
    \item The advanced expansions are designed; 1) input re-evaluation and 2) decomposition. NFCL can be enhanced by employing the input re-evaluation function $h$ \cite{nam}. To justify the structure of the input mapping function $h$, we conduct empirical analyses using real-world datasets, comparing scenarios with and without the utilization of $h$. Additionally, we perform experiments applying the average pooling method to decompose the time-series.
    \item Three versions of NFCLs are evaluated on 15 datasets collected from four diverse fields using nine forecasting benchmarks. According to experiments, NFCL-C (NFCL-Complex) achieved the most competitive performance across four metrics.
    \item The input-output directly connected network structure aids users seeking evidence for forecasting one of the time-series variables at a specific time. Independent computation for each input variable clarifies which series' time step has a more significant impact on the result.
\end{itemize}

\section{Related work}

Over the past decades, statistical and algorithmic machine learning methods have been instrumental in solving MTSF problems \cite{ARIMA_eg, GBDT_eg}. 
The evolution of NN has subsequently transformed conventional approaches to MTSF, as NNs circumvent the structural limitations inherent in traditional machine learning methods \cite{adv_nn}. 
Convolutional NNs (CNNs) and transformers have demonstrated superior performance compared to simple connected networks as NN backbone techniques in various fields where machine learning has faced recent challenges \cite{tcn, vit}.

Informer \cite{informer}, based on the transformer \cite{transformer}, has introduced its own benchmark datasets (electricity transfer temperature, ETT) to the community, establishing baselines and serving as the foundation for various methods (e.g., Autoformer \cite{autoformer}, PatchTST \cite{patchtst}, etc. \cite{fedformer, pyraformer, itransformer}) to address MTSF problems. 
Transformer-derived models not only excelled at predicting long-term time-series, where traditional recurrent NN families struggle, but also performed well at predicting short-term time-series, an area where conventional recurrent NN models dominate.
However, beyond attention mechanisms, forecasting NNs often handle cross-correlation of time-series inadequately \cite{dnlinear, patchtst}. 
Even without attention, SCINet, based on CNN, has achieved competitive performance in benchmarks for MTSF problems \cite{scinet, informer}. 
Additionally, built on simple linear weight combinations, decomposed- and normalized-linear (D-, N-Linear) models have demonstrated the best performance and interpretability simultaneously after training \cite{dnlinear}.
TimesNet \cite{timesnet} and TiDE \cite{tide} leverage CNNs and linear combinations in the model, respectively, as they pass through the input to output, achieving improved performance.

Another significant focus of our manuscript is interpretability. 
One of the reasons Attention mechanisms have garnered interest is their ability to show evidence of a model's predictions, indicating how much each point contributes during the forwarding of inputs to outputs \cite{informer}.
However, while the attention mechanism clearly indicates how much each time-series point contributes to the prediction of other time-series points, the contribution between different time-series corresponding to the input channels is not fully understood.
TimeSHAP \cite{timeshap}, a time-series explaining method utilizing SHAP \cite{shap}, summarizes contributions to simple weights for any method but requires a post-hoc process to fit the explaining model after training the recurrent NN-based forecaster. 
Nevertheless, some papers have proposed applying a simple additive structure, which can also provide contributions to model decisions but have predominantly focused on tabular datasets \cite{nam, nbm}.

\section{Preliminaries}
\subsection{Notations}

In dataset $\mathcal{D}$, each time-series is separated to two pairwise continuous series; historical and forecasting target series as $\{\mathcal{X}, \mathcal{Y}\}$,  respectively. 
The sample $X$ of historical series $\mathcal{X}$ is as,
$$X = \{ x^{1, t}, x^{2, t}, \cdots, x^{K, t} \}_{t = 1}^{L} \in \mathbb{R}^{K,L},$$
where $L$ is look-back window length and $x^{i,j}$ denote the $i$-th series at the $j$-th time step. 
Also, $K$ denotes the number of the series.
As similarly, the sample $Y$ of forecasting series $\mathcal{Y}$ is as,
$$Y=\{ x^{1, t}, x^{2, t}, \cdots, x^{K, t} \}_{t = L+1}^{L+T} \in \mathbb{R}^{K,T},$$ 
where the $T$ is forecasting time length.
Then, we supposed to hold conditions that the $L>T>1$ and $K>1$, short-term forecasting with the MTSF in the manuscript.



\subsection{Problem statement}

We defined the problem that to solve time-series forecasting is finding the optimal parameters of forecasters as, $\argmin_{\theta}\mathcal{L}(f_\theta(\mathcal{X}), \mathcal{Y})$,
where $\theta$ is the parameter of forecasting model $f_\theta: \mathbb{R}^{K,L} \rightarrow \mathbb{R}^{K,T}$ and $\mathcal{L}$ is loss function to optimize the parameter. 
Moreover, we have two kinds of datasets $\mathcal{D}$ and $\mathcal{D}'$. 
$\mathcal{D}'$ denotes the unseen dataset that we can observe in near futures and has the same shape as $L$, $T$, and $K$. 
We design the problem that the forecasting model should train and validate using only $\mathcal{D}$, and evaluate the model performance with $\mathcal{D}'$ that is excepted in the learning and validation process.

\section{NFCLs}\label{NFCL}

\subsection{Simple Expansion}

\begin{figure}[t]
  \centering
  \includegraphics[width=\columnwidth]{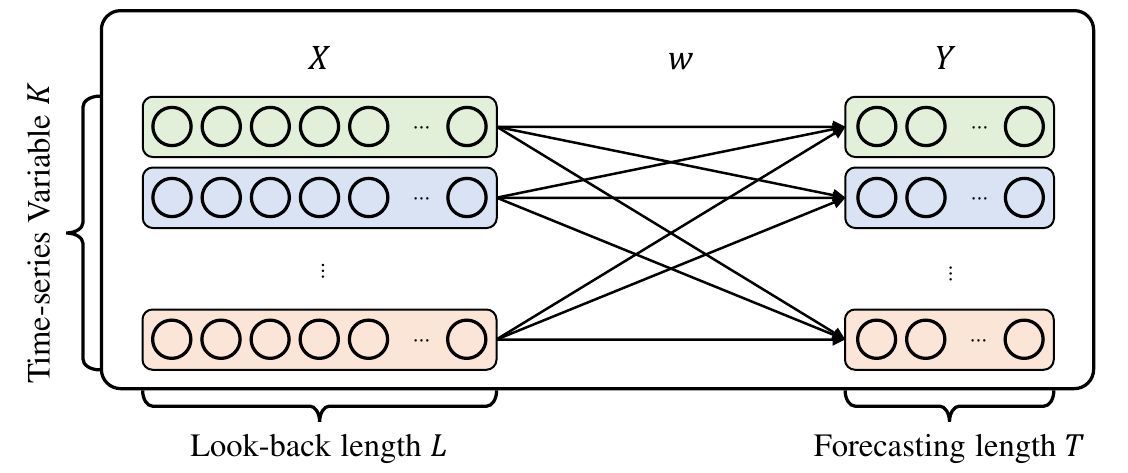}
  \caption{The simple expansion for forecasting. The black arrow denotes a fully-connected operation between each input sequence in X and each target sequence in Y.}
  \label{fig.simple_arch}
  \Description{   }
\end{figure}

\begin{figure}[t] 
  \centering
  \includegraphics[width=.99\columnwidth]{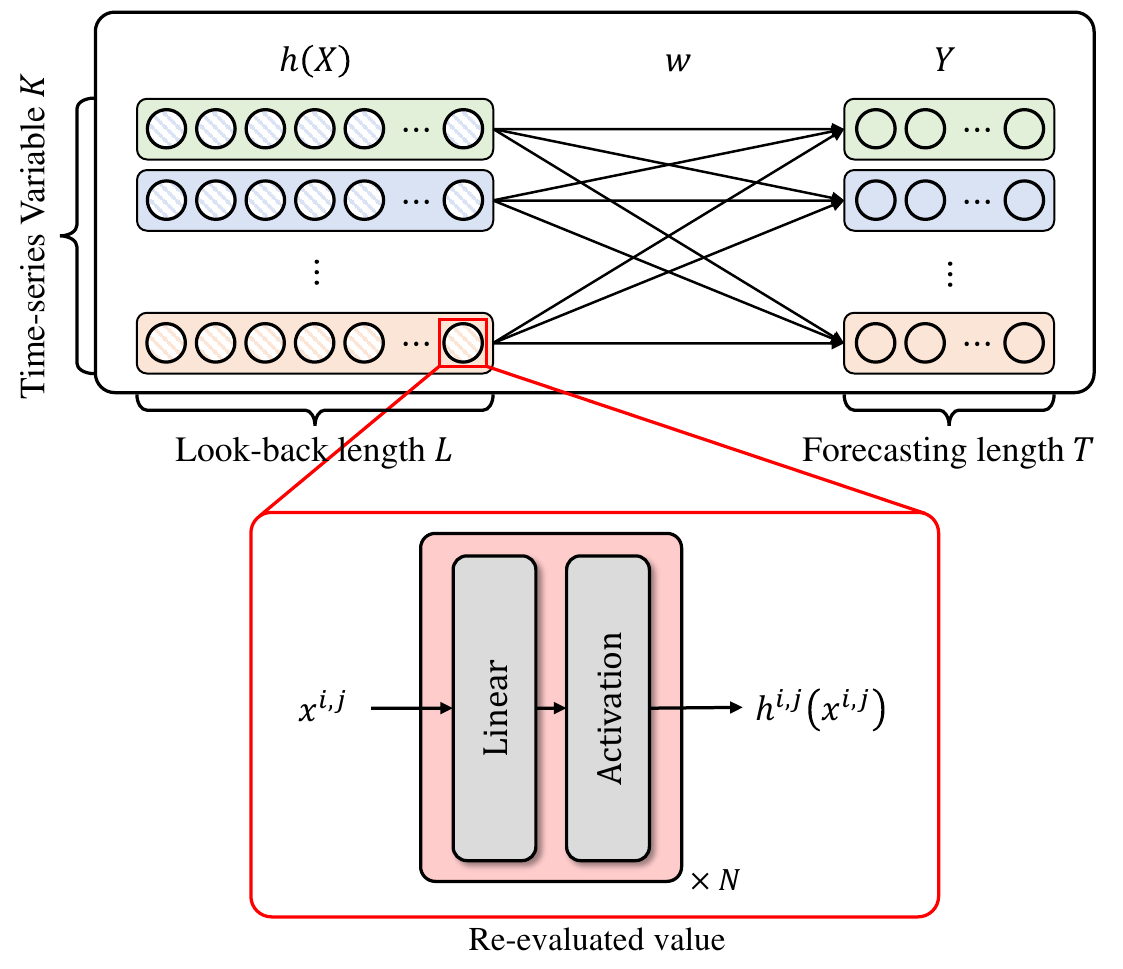}
  \caption{The complex expansion for forecasting.}
  \label{fig.complex_arch}
  \Description{  }
\end{figure}

D- and N-Linear, proposed from LSTF-Linear \cite{dnlinear}, offer an embarrassingly simple baseline for solving MTSF. 
Two linears handle the time-series as univariate to forecast the future with only one weight $w$ to compute $Xw+b$. 
For multivariate forecasting, they repeat the process of univariate as much as the number of time-series variables.
So, LSTF-Linear advocated methods that utilize capacity up to $K \cdot L \cdot T$, where ($\cdot$) represents a product between two more numbers. 
The critique suggests that LSTF-Linear neglected their approach, which improved upon using not only a univariate solution but also cross-correlated multivariate series. 
Therefore, we conducted experiments to expand the weight of the linear combinations, referred to as NFCL-V (NFCL-Vanilla).

We expanded the weight size from LSTF-Linear to $(K \cdot L) \cdot (K \cdot T) = K^2 \cdot L \cdot T$, as NFCL-V forecasts each input point (denoted as all $x^{k,t}$ in $X$) to each forecasting point (denoted as all $y^{k,t}$ in the paired $Y$). We summarized NFCL-V in Equation \ref{eq:simple_ours}:
\begin{equation}\label{eq:simple_ours}
    \hat{y}^{k,t} = \sum_{i=1}^{K}{\sum_{j=1}^{L}{x^{i,j}w^{i,j,k,t}}} + b^{k,t},
\end{equation}
where $w\in \mathbb{R}^{K,L,K,T}$ and $b \in \mathbb{R}^{K,T}$ represent the weight and bias, respectively.

Figure \ref{fig.simple_arch} depicts the process of NFCL-V. Concerning the weight $w$, the left side represents the historical input $X$, and the right side represents the forecasting target $Y$. Each circle and colored row indicates each point of each series and the time-series variable. The number of circles for each block indicates the length of $X$ or $Y$, and the arrow indicates a complete connection between the circle in the colored $X$ block and the circle in the opposite $Y$ block. The upper red box provides more details; all arrow connections between blocks in the lower box denote the complete connection between all points in blocks.

To implement the weight combination, we transformed the shape of the two matrices, weight, and bias, as $w\in \mathbb{R}^{K\cdot L, K\cdot T}$ and $b \in \mathbb{R}^{K\cdot T}$ to compute $Xw+b$. To maintain the setup, we also transformed $X$ and $Y$ into the flatten shape, as $X \in \mathbb{R}^{K\cdot L}$ and $Y \in \mathbb{R}^{K \cdot T}$.

This section introduces a method to enhance the model by adding a mapping function $h$ to the input variables. 
Several papers have explored the complication of these additive operations \cite{nam, nbm}. 
To make the model more complex, we utilize mapping methods $h$ on inputs to increase the complexity of simple connections, denoted as NFCL-C.
In Equation \ref{eq:mapping_ours}, $h : \mathbb{R}^{K\cdot L} \rightarrow \mathbb{R}^{K\cdot L}$ is an arbitrary non-linear function that maps each input $x^{i,j}$ as:

\subsection{Complex Expansion}

\begin{equation}\label{eq:mapping_ours}
    \hat{y}^{k,t} = \sum_{i=1}^{K}{\sum_{j=1}^{L}{h^{i,j}\left(x^{i,j}\right)w^{i,j,k,t}}} + b^{k,t}.
\end{equation}

Figure \ref{fig.complex_arch} depicts the process of NFCL-C. 
Most components in Figure \ref{fig.complex_arch} share the same semantics as their counterparts in Figure \ref{fig.simple_arch}. 
However, the diagonal inner patterned circle in the small red box denotes the re-evaluated value based on the input $x^{i,j}$. 
This indicates that we need $K \cdot L$ different $h^{i,j}$ to dedicate each $x^{i,j}$. 
To accelerate loop computation, we use group convolution that supports point-wise convolution, requiring the group size to be set to $K\cdot L$ \cite{depthwisecv, nbm}.
Appendix \ref{app:nfcl-implementation} illustrates the algorithm for implementing Equation \ref{eq:mapping_ours} and the corresponding algorithm utilizing group convolution.

We explored the structure of $h^{i,j}$ in experiments. 
We prepared more complex structures using not only weight combinations and non-linear activation but also dropout or batch normalization, known as NN performance-improving methods. 
However, the drop-out and normalization layer hinders the production of the re-evalu-ation value from $X$ according to someones works; hence, we conducted a simple mapping function $h$. 
Despite the uncomplicated shape of the function, the simple $h$ form of the network outperformed in most cases. 
We discuss our experimental results in Section \ref{sec:compare_hyper} to choose the proper hyperparameters for $h$.

\begin{figure}[!h]
  \centering
  \includegraphics[width=.9\columnwidth]{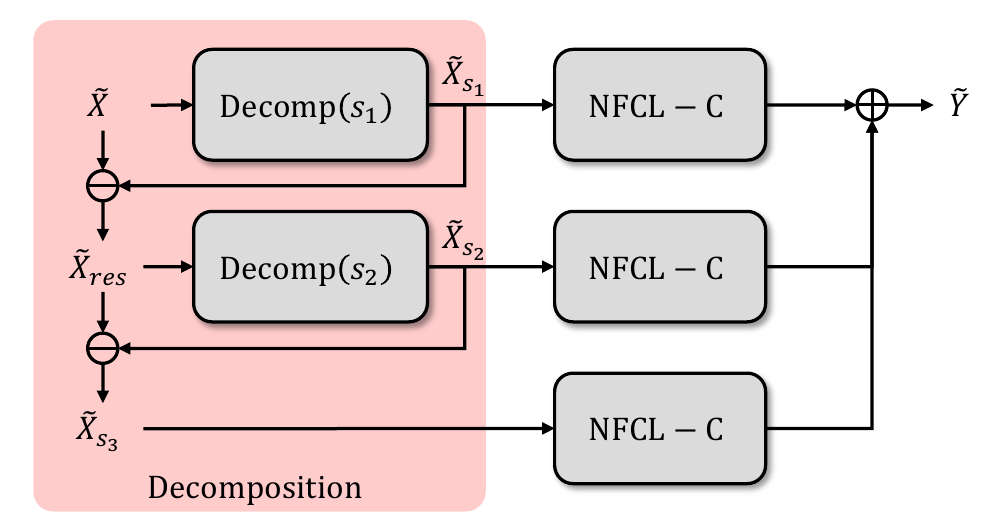}
  \caption{The decomposition for forecasting.}
  \label{fig.decomp_nfcl}
  \Description{  }
\end{figure}

\begin{figure}[!htp]
  \centering
  \includegraphics[width=\columnwidth]{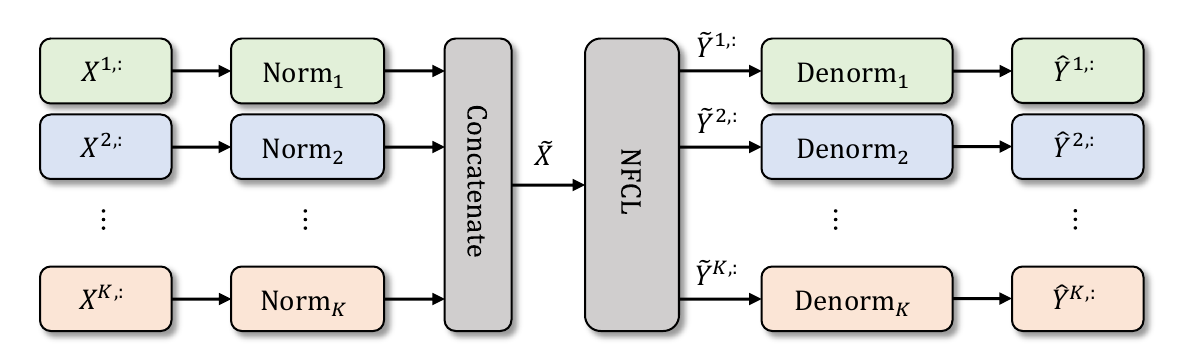}
  \caption{The NFCLs overall pipeline for forecasting.}
  \label{fig.whole_arch}
  \Description{  }
\end{figure}

\subsection{Decomposed Expansion}

On one hand, we consider pre-processing, such as time-series decomposition. 
Several papers have focused on forecasting improvement based on average pooling and frequency analysis \cite{autoformer, dnlinear}. 
Thus, we prepared the separation for time-series with two kinds of methods: 1) decomposition and 2) statistical normalizing.

The first is moving average pooling with the kernel size $S$ that has three over diverse. 
We conduct the pooling sequence from the largest $s_1$ to the smallest $s_N$. 
Before applying the pooling sequence, the time axis of $\tilde{X}$ is padded to the left side with zero to keep the time-series length as $L$. 
At the end of each pooling sequence, $\tilde{X}$ is separated into two over decomposed series $\tilde{X}_{dec}=[\tilde{X}_{s_1}, \cdots, \tilde{X}_{s_N}]$, where $S = \left\{s_1,\cdots, s_N ; s_{n} > s_{n+1}\right\}$. 
Note that the smallest $s_N$ must be $1$ to capture the residual used to recover the raw time-series, while the largest $s_1$ shows the trend of time-series. 
We built independent models to dedicate each $\tilde{X}_{s_{n}}$ respectively and summed all outputs of each $h_s$. 
In the manuscript, the decomposed NFCL is denoted as NFCL-D (NFCL-Decomposition), and details are followed in appendix \ref{app:decposition_algorithm}.
Figure \ref{fig.decomp_nfcl} illustrates the time-series decomposition flow. 
Within the red area in Figure \ref{fig.decomp_nfcl}, the time-series $\tilde{X}$ is decomposed into Decomp. blocks. 
Each decomposition block comprises time-series left padding and average pooling. 
Please refer to Appendix \ref{app:decposition_algorithm} for more details.

Next is statistical normalizing $X$ according to each time-series variable. 
In this case, the normalized input $\tilde{X}$ is delivered to the arbitrary function $h$. 
After extraction from $h$, the mapping value $h(\tilde{X})$ is recovered as much as the shifted values from applying normalization. 
This separation makes the module work easy: the first module's role is only normalization, and the second module's role is to capture the forecasting pattern.
We summarized the process with Equation \ref{eq:norm_eq} as follows:
\begin{align}\label{eq:norm_eq}
\begin{split}
    \tilde{x}^{k,:} &= \alpha \left( {\frac{x^{k,:} - \mathrm{mean}(x^{k,:})}{\mathrm{std}(x^{k,:})}} \right) + \beta,  \\
    \hat{y}^{k,:} &= \frac{\tilde{y}^{k,:} - \beta}{\alpha} \cdot \mathrm{std}(x^{k,:}) + \mathrm{mean}(x^{k,:}),
\end{split}
\end{align}
where $(:)$, $\mathrm{mean}$s, and $\mathrm{std}$s indicate all times of $k$ time-series, the functions of average and standard deviation, respectively. 
Moreover, $\alpha\in \mathbb{R}^{k}$ and $\beta\in \mathbb{R}^{k}$ are the learnable parameters that apply normalization effectively \cite{layernorm}. 
In short, input $X$ is normalized to $\tilde{X}$, $h$ extracts the contribution based on the center-adjusted time-series $\tilde{X}$, and normalized target $\tilde{Y}=h(\tilde{X})w+b$ is invert-shifted to $\hat{Y}$ with the original $X$ statistics.
Figure \ref{fig.whole_arch} illustrates the overall pipeline of NFCLs for the forecasting process.
Each time-series that is colored unique in a historical time-series $X$ is subjected to time-series specific normalization as $\tilde{X}$ shown in Equation \ref{eq:norm_eq}.
$\tilde{X}$ are concatenated along the first dimension and passed to the NFCLs.
NFCLs predict the $\tilde{Y}$s from the normalized input $\tilde{X}$.
It then applies denormalization to each time-series, reusing the statistical values for each time-series used in the normalization.
The denormalization is inversely applied to the $\tilde{Y}$s, resulting in $Y$.





\section{experiments}\label{sec:experiments}

\begin{table*}[!tp]
\centering
\resizebox{.9\textwidth}{!}{%
\begin{tabular}{cccccccccccc} \toprule \toprule
\multicolumn{1}{l}{Metric} & Category    & NFCL                                   & DLinear & NLinear     & TiDE                                  & Informer & Autoformer & PatchTST                               & iTransformer                           & SCINet                                & TimesNet     \\  \cmidrule(lr){1-2} \cmidrule(lr){3-12} \morecmidrules \cmidrule(lr){1-2} \cmidrule(lr){3-12}
                           & Patient     & 1.212                                  & 1.140   & 1.072       & 1.016                                 & 1.362    & 1.142      & 0.939                                  & {\color[HTML]{0000FF} \textbf{0.781}}  & 1.145                                 & \ul{0.864}  \\
                           & Currency    & 0.095                                  & 0.090   & \ul{0.090} & {\color[HTML]{0000FF} \textbf{0.090}} & 0.761    & 0.121      & 0.090                                  & 0.094                                  & 0.222                                 & 0.102        \\
                           & Electricity & {\color[HTML]{0000FF} \textbf{0.299}}  & 0.308   & 0.311       & 0.311                                 & 0.375    & 0.349      & \ul{0.299}                            & 0.311                                  & 0.305                                 & 0.300        \\
\multirow{-4}{*}{MAE}      & Climate     & {\color[HTML]{0000FF} \textbf{0.310}}  & 0.361   & 0.348       & 0.346                                 & 0.383    & 0.391      & 0.334                                  & 0.340                                  & \ul{0.322}                           & 0.328        \\ \cmidrule(lr){1-2} \cmidrule(lr){3-12}
                           & Patient     & 6.349                                  & 3.080   & 2.843       & \ul{2.698}                           & 4.312    & 3.018      & 2.972                                  & {\color[HTML]{0000FF} \textbf{2.326}}  & 3.238                                 & 3.413        \\
                           & Currency    & 0.023                                  & 0.021   & 0.021       & \ul{0.021}                           & 0.975    & 0.031      & {\color[HTML]{0000FF} \textbf{0.021}}  & 0.022                                  & 0.110                                 & 0.025        \\
                           & Electricity & 0.245                                  & 0.249   & 0.266       & 0.265                                 & 0.304    & 0.292      & 0.245                                  & 0.257                                  & {\color[HTML]{0000FF} \textbf{0.231}} & \ul{0.241}  \\
\multirow{-4}{*}{MSE}      & Climate     & \ul{0.314}                            & 0.354   & 0.373       & 0.370                                 & 0.394    & 0.400      & 0.355                                  & 0.355                                  & {\color[HTML]{0000FF} \textbf{0.314}} & 0.336        \\ \cmidrule(lr){1-2} \cmidrule(lr){3-12}
                           & Patient     & 38.149                                 & 43.689  & 41.671      & 40.446                                & 52.267   & 43.967     & 34.421                                 & {\color[HTML]{0000FF} \textbf{30.707}} & 43.916                                & \ul{32.280} \\
                           & Currency    & 7.066                                  & 6.820   & \ul{6.816} & {\color[HTML]{0000FF} \textbf{6.800}} & 34.877   & 8.835      & 6.857                                  & 7.047                                  & 12.000                                & 7.499        \\
                           & Electricity & \ul{27.546}                           & 28.366  & 28.140      & 28.113                                & 34.892   & 32.096     & {\color[HTML]{0000FF} \textbf{27.532}} & 28.471                                 & 29.137                                & 28.100       \\
\multirow{-4}{*}{SMAPE}    & Climate     & {\color[HTML]{0000FF} \textbf{29.043}} & 33.533  & 31.415      & 31.222                                & 35.814   & 36.123     & 30.494                                 & 31.039                                 & 31.460                                & \ul{30.419} \\ \cmidrule(lr){1-2} \cmidrule(lr){3-12}
                           & Patient     & -0.484                                 & -0.179  & 0.269       & \ul{0.274}                           & -0.580   & 0.244      & 0.074                                  & {\color[HTML]{0000FF} \textbf{0.339}}  & -0.050                                & 0.144        \\
                           & Currency    & 0.931                                  & 0.928   & 0.929       & 0.929                                 & -2.475   & 0.912      & {\color[HTML]{0000FF} \textbf{0.936}}  & \ul{0.932}                            & 0.603                                 & 0.929        \\
                           & Electricity & {\color[HTML]{0000FF} \textbf{0.650}}  & 0.636   & 0.636       & 0.635                                 & 0.350    & 0.565      & 0.644                                  & 0.623                                  & 0.617                                 & \ul{0.644}  \\
\multirow{-4}{*}{$R^2$}    & Climate     & {\color[HTML]{0000FF} \textbf{0.659}}  & 0.614   & 0.595       & 0.600                                 & 0.366    & 0.464      & 0.616                                  & 0.611                                  & \ul{0.639}                           & 0.633       \\ \bottomrule \bottomrule
\end{tabular}%
}
\caption{The average score for each model by metric per dataset category. Each score is an average of the scores across datasets in each category and forecast length $L$. }
\label{tab:category_scores}
\end{table*}

\subsection{Datasets}


We prepared 15 datasets collected from various sources, including {\bf Patient}, {\bf Currency}, {\bf Electricity}, and {\bf Climate}. 
For testing the MTSF problem, all $K$ time-series variables were utilized during both training and evaluation. 
We adopted the evaluation concept of Informer, dividing all datasets into training, validation, and test sets in a ratio of 6:2:2 for all datasets \cite{informer}. 
Meanwhile, the Informer employed a 7:1:2 splitting ratio exclusively for ETT datasets, while others utilized a 6:2:2 splitting ratio. 
We adjusted Informer's evaluation procedure 7:1:2 split to 6:2:2 to maintain consistency in the splitting approach.
Experiments were conducted with different length settings than those of Informer, setting the historical length $L$ to $24$ and the predicted length $T \in \{6,12,18,24\}$. 
It's important to note that we strictly separated $\mathcal{D}$ and $\mathcal{D}'$ during both the training and testing phases. 
Additional details, including dataset abbreviations, are provided in Appendix \ref{app:detail_dataset}.

\subsection{Metrics}
The employed metrics in our experiment include four widely recognized measures: Mean Absolute Error (MAE), Mean Squared Error (MSE), Symmetric Mean Absolute Percentage Error (SMAPE), and \(R^2\). 
Through our experiments, we observed that the popular non-scaled metric tended to overestimate the baselines. 
While MAE and MSE portray non-scaled errors, SMAPE and \(R^2\), which are bounded between 0 and 1 or percentage, facilitate the quantification of errors across diverse scales. 
Additionally, SMAPE and \(R^2\) are related to MAE and \(R^2\), respectively. 
$R^2$ is another widely used metric to evaluate regression results, as it expresses the explained variances in $Y$ by a model related to the entire variance of $Y$.
Notably, SMAPE proves to be a superior metric compared to Mean Absolute Percentage Error (MAPE). 
SMAPE addresses the issue of MAPE diverging to \(\infty\) when \(Y \simeq 0\) and \(0 < Y \ll \hat{Y}\) \cite{smape}.
For comprehensive details on the metrics, please refer to Appendix \ref{app:metrics}.

\subsection{Baseline}

The nine baselines are categorized into three types of forecasters: 1) fully connected, 2) CNN-based, and 3) attention-based models. 
So, we employed models as; \bd{D-Lienar}, \bd{N-Linear}, \bd{TiDE}, \bd{Informer}, \bd{Autoformer}, \bd{PatchTST}, \bd{iTransformer}, \bd{SCINet}, and \bd{TimesNet}.
Note that we did not derive our experimental results from another paper. 
Instead, we independently re-implemented each paper's methodology and hyperparameter setting based on their descriptions and GitHub repositories. 
Subsequently, we rigorously re-tested these implementations on our dataset split.
Although we considered testing additional baselines, a few models did not work as expected, possibly due to issues with the official repositories on GitHub.
Moreover, some others could work for forecasting while the models perform too much poorly.
A brief abstraction of each baseline's properties is attached on Appendix \ref{app:hyper_params}.


\begin{table*}[th]
\centering
\resizebox{.9\textwidth}{!}{%
\begin{tabular}{cccccccccccc} \toprule \toprule
Metric                  & Type & NFCL                                 & DLinear & NLinear & TiDE    & Informer & Autoformer & PatchTST     & iTransformer & SCINet                             & TimesNet   \\\cmidrule(lr){1-2} \cmidrule(lr){3-12} \morecmidrules \cmidrule(lr){1-2} \cmidrule(lr){3-12}
                        & Rank & {\color[HTML]{0000FF} \textbf{2.33}} & 6.98    & 6.05    & 5.33    & 9.13     & 9.25 & 3.58       & 4.95    & 3.93                               & \ul{3.45} \\
\multirow{-2}{*}{MAE}   & Win  & {\color[HTML]{0000FF} \textbf{43}}   & 3       & 0       & 4       & 0        & 0    & 1          & 4       & \ul{5}                            & 0          \\\cmidrule(lr){1-2} \cmidrule(lr){3-12}
                        & Rank & {\color[HTML]{0000FF} \textbf{2.85}} & 5.08    & 6.78    & 6.10    & 8.52     & 9.03 & 4.45       & 5.08    & \ul{3.15}                         & 3.95       \\
\multirow{-2}{*}{MSE}   & Win  & \ul{23}                             & 3       & 0       & 0       & 0        & 0    & 5          & 4       & {\color[HTML]{0000FF} \textbf{25}} & 0          \\\cmidrule(lr){1-2} \cmidrule(lr){3-12}
                        & Rank & {\color[HTML]{0000FF} \textbf{2.00}} & 6.95    & 5.70    & 4.82    & 9.28     & 9.30 & \ul{3.08} & 4.72    & 5.68                               & 3.47       \\
\multirow{-2}{*}{SMAPE} & Win  & {\color[HTML]{0000FF} \textbf{43}}   & 0       & 3       & \ul{4} & 0        & 0    & 3          & \ul{4} & 0                                  & 3          \\ \cmidrule(lr){1-2} \cmidrule(lr){3-12}
                        & Rank & {\color[HTML]{0000FF} \textbf{2.30}} & 5.32    & 6.40    & 5.87    & 9.05     & 8.98 & 4.40       & 4.98    & 4.07                               & \ul{3.63} \\ 
\multirow{-2}{*}{$R^2$} & Win  & {\color[HTML]{0000FF} \textbf{37}}   & 0       & 4       & 0       & 0        & 2    & 5          & 2       & \ul{10}                           & 0         \\ \bottomrule \bottomrule
\end{tabular}%
}
\caption{The average rank and number of wins of performance on experiments versus benchmarks.}
\label{tab:summary_benchmark}
\end{table*}

\subsection{Experiments Setting}

Our experiments were implemented using PyTorch 2.X \cite{pytorch}. 
While most forecasting papers have followed the experimental code of Autoformer, we structured our experiments based on PyTorch-Lightning for improved reusability among researchers.
Our GitHub will provide the source codes to develop forecasting applications for handling time-series datasets.

We conducted multiple runs of our experiments, with each model learning and testing each dataset up to five times using different seeds and a batch size of 128. 
To optimize the models, we employed the AdamW optimizer \cite{adamw} with a weight decay $\lambda$, and the learning rate $\mathrm{lr}$ was set to 0.001. 
We aimed for sufficient convergence, utilizing 100 early-stopping rounds with MSE as the criterion. 
The model's trainable weights were reset to their best-performing state at the end of the training process.
The recovered model was then employed in the testing phase using the dataset $\mathcal{D}'$.

\subsection{NFCL Setting}

All three versions of NFCL underwent validation on the test bed with uniform normalization applied across all variants. 
The results indicated that NFCL-C with a single layer employing 32 hidden nodes demonstrated the best performance. 
Consequently, NFCL-C was used in Subsection \ref{sec:compare_all} to represent NFCL. 
In addition, NFCL-C employed the leaky-ReLU activation. 
There is no special reason why we choose the leaky-ReLU that was made for its ability to express positive and negative values (not close to 0) non-linearly.
Furthermore, we have not considered the task of forecasting univariate time-series, as NFCL has a similar concept to LSTF-linear when dealing with univariate time-series \cite{dnlinear}.

We note that each version of NFCL was designated as follows: Simple Forward (NFCL-V), Using Mapping Function $h$ (NFCL-C), and Applied Decomposition (NFCL-D). 
In the investigation, NFCL-C and -D were examined with diversity, incorporating four different numbers of hidden nodes, three distinct layer numbers, and variations in the application of time-series decomposition.
The term NFCL, without any suffix, refers to the best-performing NFCL-C, as we avoided comparing NFCLs across experiments to prevent overestimation. 
Thus, NFCL-C was compared to other versions in experiments, using the full names of each NFCL in additional experiments.

\subsection{NFCL's Performances}\label{sec:compare_all}

Here are Table \ref{tab:category_scores} and \ref{tab:summary_benchmark} displaying the performance of NFCL and the baselines.
Table \ref{tab:category_scores} and \ref{tab:summary_benchmark} summarizes the results based on the information provided in appendix \ref{baseline-comparison}, that includes the full benchmark results according to each metric.
In the scores of the tables, \bd{blued-bold} and \ul{underlined} scores indicate the best and next best scores, respectively.
In addition, Table \ref{tab:category_scores} has some similar scores not highlighted because they have differences under the four decimal points.
Note that NFCL in the column refers to NFCL-C with $h = 32$ nodes and one layer.

Table \ref{tab:category_scores} averaged the scores depending on the category where the dataset is included and the forecast length $L$.
The first and second columns in Table \ref{tab:category_scores} show the metric and the category, respectively (Appendix \ref{app:detail_dataset} notice you to which category included each dataset). 
Based on the aggregate results, our NFCL performed best on Electricity and Climate categories when utilizing metrics other than MSE. 
While NFCL performed worse than the best-performing reference in terms of MSE, NFCL excelled in $R^2$, the normalized score of MSE. 
Whereas MSE indicates that NFCL is performing worse than the other references for the squared deviation of $Y$ from $\hat{Y}$, $R^2$ shows that NFCL is performing better than the other benchmarks for the predicted deviation from the variance of $Y$. 
Therefore, NFCL demonstrates its ability to capture the variability in the dataset better than other benchmarks when the real-world data has a large variance. 
Additionally, even on datasets in the Currency, NFCL did not show significant degradation compared to other benchmarks. 
However, NFCL performed poorly in the Patient, which we attribute to overfitting of NFCL due to the small number of data samples (under 600 training points, and 200 validation and test points).

Table \ref{tab:summary_benchmark} summarizes the results based on the scores provided in appendix \ref{baseline-comparison}, which includes the full benchmark results according to each metric.
The first and second columns in each table show the metrics and the counter type that evaluated the performance between real $Y$ and forecasted $\hat{Y}$, while the third column onwards points to the respective average ranks or the number of wins depend on the model utilized. 
In Table \ref{tab:summary_benchmark}, concerning MAE, NFCL demonstrates an average ranking of 2.33, and it outperforms competitors in 43 out of 60 total testbeds. 
Regarding MSE, NFCL achieves the best performance 23 times, which is two instances fewer than SCINet; however, NFCL maintains a superior average ranking. 
Additionally, NFCL excels in scaled metrics, such as SMAPE and $R^2$. 
As indicated in Table \ref{tab:summary_benchmark}, the proposed model attains an average ranking of 2.00 and 2.30 for SMAPE and $R^2$, respectively, securing the top rank in 43 and 37 instances.

There is a discernible contrast in the performance of the models with and without scaling, with the most recently studied baselines exhibiting relatively superior results on scaled metrics. 
Informer and Autoformer performed inadequately for short time-series forecasting in all cases. 
SCINet exhibited a significant decline in performance on scaled metrics compared to its performance before scaling, whereas PatchTST and TimesNet demonstrated the opposite behavior. 
TiDE, a fully connection-based approach, outperformed D- and N-Linear in most cases but did not surpass the performance of the other baselines.

We also investigated the learnable parameters for NFCL-V and NFCL-C with benchmarks, which are presented in Appendix \ref{app:count_params}.
According to Table \ref{tab:detail_data}, the number of multivariate time-series in the dataset we tested is $K \in \left\{21, 10, 8, 7\right\}$. 
The number of learning parameters as a function of the number of time-series and forecast length variation was larger for the transformer series benchmark than for the other references. 
The smallest models are LSTF-Linear, with parameters $L \cdot T$. 
The model with the next smallest number of learning parameters was SCINet. 
The next model that consumes more capacity than SCINet is NFCL-V, followed by NFCL-C.

\subsection{Comparison of NFCL's module}\label{sec:compare_hyper}

In this section, we provide a summary of the hyperparameter search process for NFCL, as detailed in the preceding section. 
For the purpose of experiment comparison, we computed the average ranking for each of the four metrics, as presented in the summary of the complete experiment table among only the variation of NFCLs, not including the references. 
Table \ref{tab:hyper_comparison} illustrates the best performer election process of NFCL across the experiment pipeline.
The first row of Table \ref{tab:hyper_comparison} delineates the categories for the subsequent rows. 
Accordingly, the first and second columns represent the variations of NFCL and the structure of $h$. 
If you observe (-) in $h$ in Table \ref{tab:hyper_comparison}, it corresponds to NFCL-V (in case there is no $h$). 
Within the list in $h$, the numbers denote the count of hidden nodes in each layer, with multiple numbers indicating stacked layers. 
Notably, we employed $s_n \in \{10, 4, 1\}$ as the decomposition kernel size for NFCL-D. 
For each decomposed time-series, NFCL-C was configured, and the dimension of $h$ utilized in NFCL-D matches the description in NFCL-C. 
Similarly, the best scores in Table 3 are presented in bold, and the next best scores are underlined. 
The full tables of the NFCL's hyperparameter comparison are recorded in Appendix \ref{app:nfcm_compare}.

The number of possible combinations of $h$ and decomposition structures in NFCL's experiments is vast. 
The experiments to optimize the structure for 15 datasets were conducted in the following order:
\begin{enumerate}
    \item We investigated whether NFCL-C is more effective than NFCL-V. 
We initially tried $h \in \{[64], [64, 64]\}$ and found that NFCL-C outperforms NFCL-V on all metrics except SMAPE.
    \item We assessed whether to utilize decomposition or not to NFCL. 
Based on experiments with one or two layers, we found that attempting decomposed MTSF resulted in worse performance in all cases.
    \item We explored the depth structure of $h$ that performed best on NFCL-C. 
We went as deep as possible, utilizing combinations of 64 and 128 hidden nodes. 
Experiments showed that performance degraded as the number of layers increased.
    \item We investigated the number of hidden nodes that worked best when only one layer was utilized in NFCL-C. 
For one layer, we found that performance dropped again when hidden nodes were set to 16 or less.
\end{enumerate}
Table \ref{tab:hyper_comparison}, on the other hand, is just an average ranking based on the tables recorded in Appendix \ref{app:nfcm_compare}. 
While the average ranking changes dramatically, we can see that there is not much difference between the actual scores.
According to Appendix \ref{app:nfcm_compare}, NFCL-V performs well on the ETTh1, and there are datasets and $T$ for which increasing the layer of $h$ is significant. 
In addition to the average rankings in Table \ref{tab:hyper_comparison}, we also looked at the number of times the model performed best as we went through the four-step hyperparameter exploration described above.
Therefore, we use the NFCL-C as NFCL utilized in this paper was set to utilize $h$ with one layer and the number of hidden nodes set to 32.
We conducted an analysis to identify the factors contributing to the underperformance of NFCL-D. 
It was observed that NFCL-D exhibited a more pronounced overfitting phenomenon compared to other NFCL variants, likely attributed to the application of decomposition. 
For further insights, please consult Appendix \ref{app:nfcl_generalization}.

\begin{table}[tp]
\centering
\resizebox{.9\columnwidth}{!}{%
\begin{tabular}{lcccccc} \toprule \toprule
Variation               & $h$          & MAE   & MSE   & SMAPE & $R^2$ & Avg. rank \\ \cmidrule(lr){1-2} \cmidrule(lr){3-6} \cmidrule(lr){7-7} \morecmidrules \cmidrule(lr){1-2} \cmidrule(lr){3-6} \cmidrule(lr){7-7}
NFCL-V                  & -            & 8.42  & 9.60  & 5.27  & 9.37  & 8.16      \\ \cmidrule(lr){1-2} \cmidrule(lr){3-6} \cmidrule(lr){7-7}
\multirow{9}{*}{NFCL-C} & 16           & 4.57  & 5.08  & 4.45  & 5.37  & 4.87      \\ 
                        & 32           & \textcolor{blue}{\bd{3.93}}  & \textcolor{blue}{\bd{4.42}}  & \textcolor{blue}{\bd{4.27}}  & \textcolor{blue}{\bd{4.65}}  & \textcolor{blue}{\bd{4.32}}      \\
                        & 64           & \ul{4.13}  & \ul{4.48}  & \ul{4.28}  & \ul{4.68}  & \ul{4.40}      \\
                        & 128          & 5.15  & 5.25  & 5.62  & 5.52  & 5.38      \\
                        & 16, 16       & 5.13  & 4.68  & 5.83  & 4.78  & 5.11      \\
                        & 32, 32       & 5.72  & 4.90  & 5.87  & 5.18  & 5.42      \\
                        & 64, 64       & 6.47  & 6.23  & 6.80  & 6.80  & 6.58      \\
                        & 128, 128     & 7.38  & 6.78  & 8.12  & 7.28  & 7.39      \\
                        & 128, 128, 64 & 7.88  & 7.05  & 9.10  & 7.50  & 7.88      \\ \cmidrule(lr){1-2} \cmidrule(lr){3-6} \cmidrule(lr){7-7}
\multirow{4}{*}{NFCL-D} & 64           & 10.45 & 10.85 & 10.28 & 10.12 & 10.43     \\
                        & 128          & 11.08 & 11.35 & 10.95 & 10.73 & 11.03     \\
                        & 64, 64       & 12.12 & 12.08 & 11.82 & 11.32 & 11.83     \\
                        & 128, 128     & 12.57 & 12.23 & 12.35 & 11.70 & 12.21     \\ \bottomrule \bottomrule
\end{tabular}%
}
\caption{Hyper-parameter analysis of the number of layers and nodes according to NFCL variations.}
\label{tab:hyper_comparison}
\end{table}

\subsection{NFCL interpretation}\label{sec:interpret}
\begin{figure}[ht]
  \centering
  \includegraphics[width=.95\columnwidth, right]{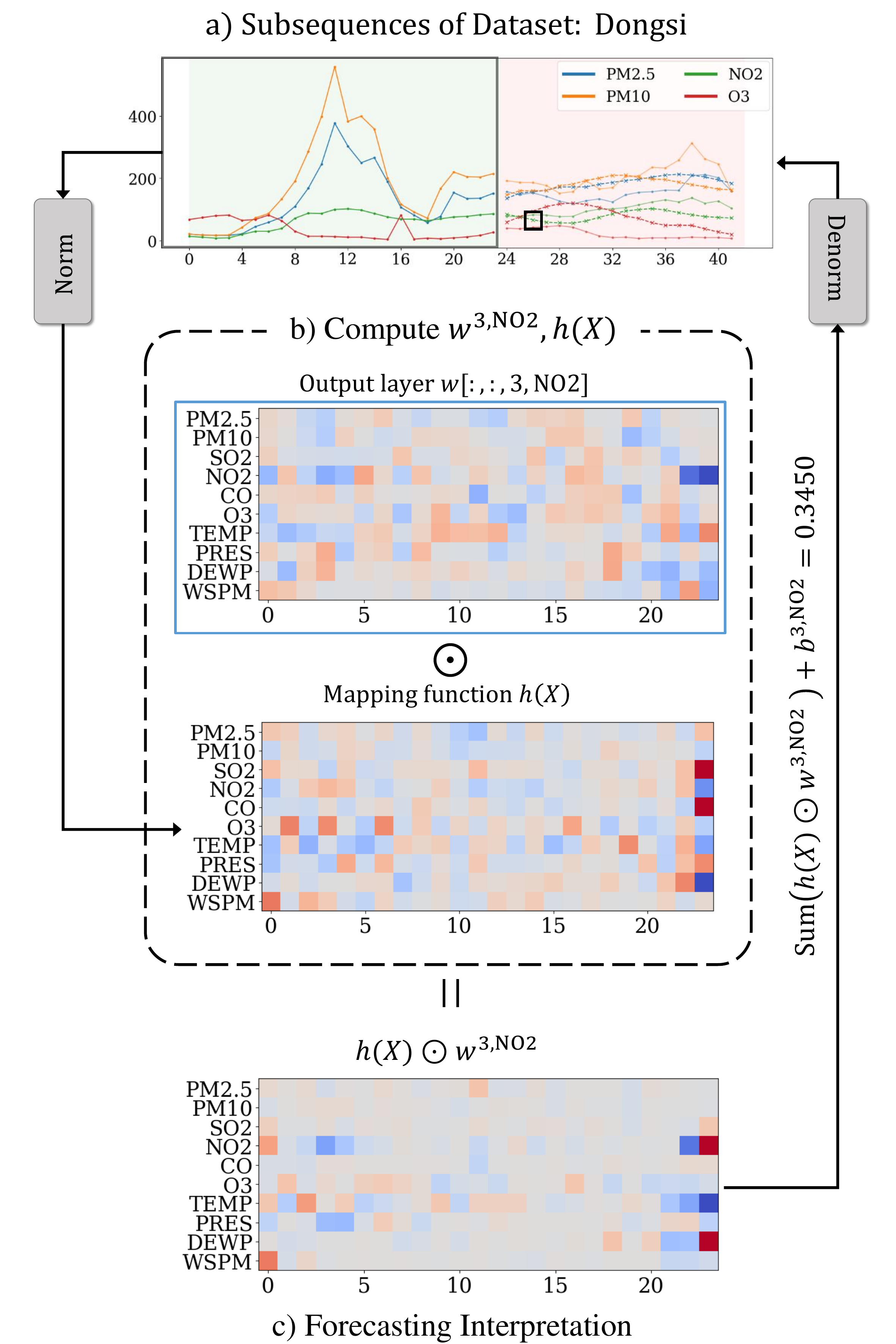}
  \caption{Illustration of NFCL Interpretations.}
  \label{fig.interpretation}
  \Description{  }
\end{figure}

After training, NFCL calculates the direct contribution for each point in $X$. 
Figure \ref{fig.interpretation} depicts the process of contribution calculating. 
The data used originates from one of Dongsi's test samples, with only a few portions selected as a visualization example for the illustration. 
The Dongsi dataset recorded the atmospheric sensor's value of two inhalable Particulate Matter (PM2.5 \& PM10, where a number is a micron radius of matter) in the air, including various climate values. 
Figure \ref{fig.interpretation} displays the data utilized in sub-plots a), b), and c).

In sub-plot a) of figure \ref{fig.interpretation}, the $X$ of the green area is used to forecast $Y$ of the red area. 
The continuous line with the circle marker and dotted line with the x-marker of sub-plot a) denotes the real and predicted series. 
Sub-plot b) shows the outcome of computing $w$, and $h(X)$ based on $X$, while sub-plot c) displays the contribution attributed to $h(X) \odot w$ in predicting the result of $y_{k,t}$. 
$X$ is input into the mapping function $h$ in the sub-plot b), where it undergoes re-evaluation. 
This result is then combined with $w$, assigning weights to contributions from all variables, producing the final contribution map. 
We execute $h(X) \odot w$ as a matrix multiplication, equivalent to the Hadamard product between the two matrices, denoted by the $\odot$ operator in the sub-plot b), followed by summation. 
Thus, the map in the sub-plot c) depicts a scenario where only $\odot$ is performed. 
By reshaping the section of $w$ highlighted by the blue or green box in the sub-plot b), conducting $h(X)$ and $\odot$, we obtain the corresponding figure in the same color in the sub-plot c).
This NFCL interpretation process can be applied to all input variables to forecast a variable at a future point in time. 
Specifically, the weight map for the output layer $w$ is excessively large and is not detailed in main context but is available in Appendix \ref{full_weight_map}.
The width-broad, and height-narrow blue box region in Figure \ref{fig.w-cdot-hx}, as illustrated in Appendix \ref{full_weight_map}, represents $w\left[:,:,3,\mathrm{NO2}\right]$.

Examining the forecast for NO$_2$ three hours into the future, although a more diverse pattern than PM2.5 and PM10 are not identified, it is observed that NO$_2$ is autocorrelated and contributed to by factors such as TEMP, DEWP, and WSPM. 
Notably, the input consists of 24 points, representing a day in hourly data. 
Consequently, the initial part of the historical time-series aligns with the same time as the first point in $Y$, indicating a robust day-to-day periodicity, particularly in climate-related data.

\subsection{Analysis contributions}\label{sec:analysis_cont}




Figure \ref{fig.ip_eg} presents the results of executing the two examples in Figure \ref{fig.w-cdot-hx} for all $w$ in Appendix \ref{full_weight_map}. 
For each subplot, a left-to-right shift on the X-axis indicates an increase in the historical $L$, and a top-to-bottom shift on the Y-axis indicates an increase in the forecasting $T$. 
The low contribution $h(x)\odot w$ is shown in grayscale, and we applied a similar approach to the other plots (we used grayscale because using a separate color depth scale for each plot would not allow us to accurately perceive the global contribution to a particular prediction).

The upper green-edged subplot illustrates the association of particulate matter with SO2 and NO2. 
Examining the summation of contributions, PM2.5 receives a larger contribution from SO2 than other variables and a smaller contribution from NO2. 
Notably, In China, research results have shown a strong correlation between PM2.5 and SO2, while NO2 exhibited inconsistent patterns, with correlations depending on the region \cite{so2andpm2}.
The autocorrelation and cross-correlation for O3, temperature, and barometric pressure forecasts are visible in the red-outlined contribution map in the lower red-edged subplot.
Indeed, O3 shows a strong correlation with surface air temperature \cite{o2_temp}, and depending on the terrain, wind can further enhance the O3–temperature correlation \cite{wind_o2_temp}.
Visualizing the contributions across time-series can be an important element in laying the groundwork for future applications and use of NFCLs by ML Non-Experts in other fields, not just the one to which this study was applied.

\begin{figure}[t]
  \centering
  \includegraphics[width=.8\columnwidth]{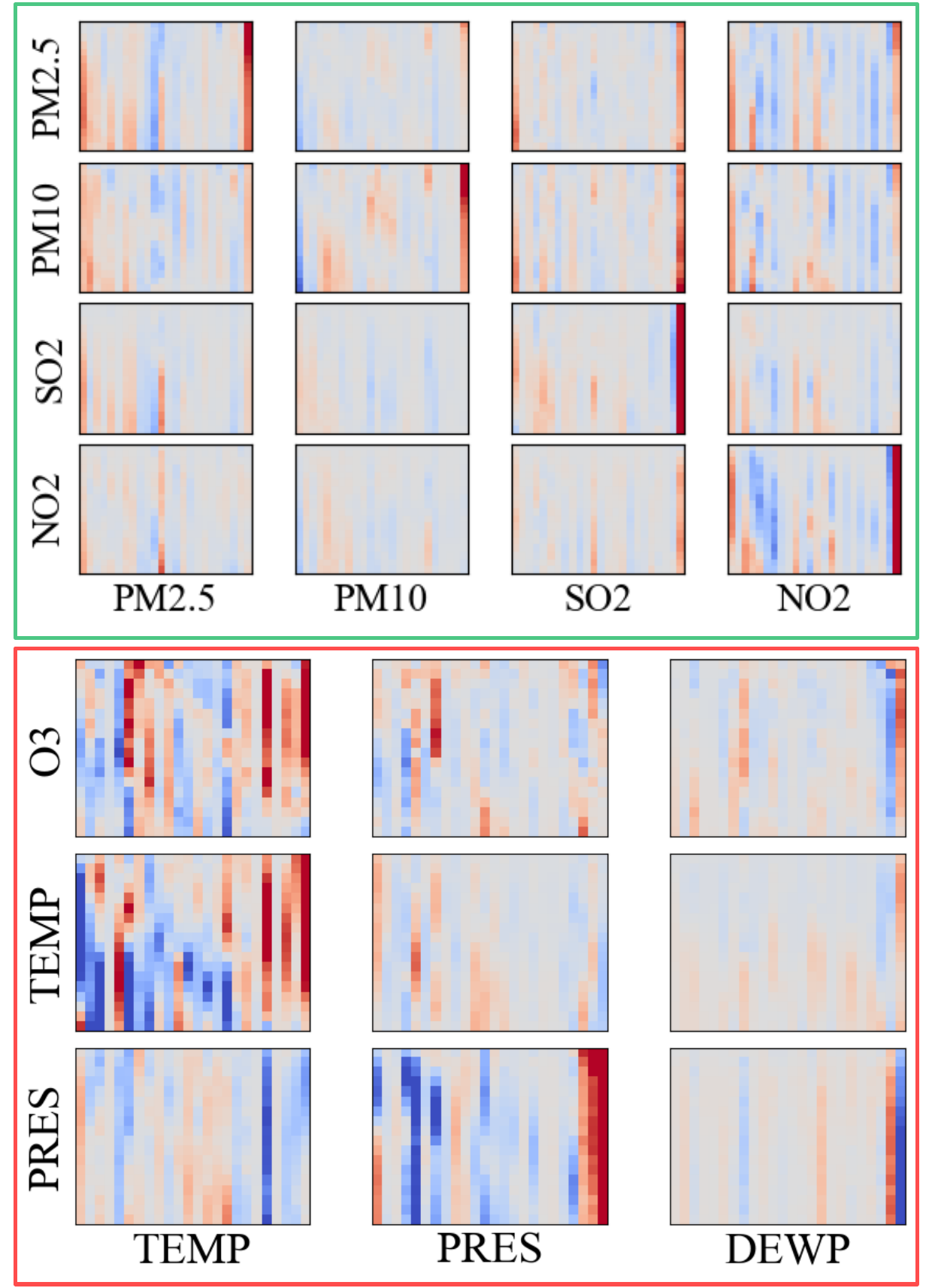}
  \caption{Interpretation example.}
  \label{fig.ip_eg}
  \Description{  }
\end{figure}

\section{Conclusion}

We discuss various variants of NFCL in our paper. 
Essentially, NFCL provides a method to leverage auto-correlation and cross-correlation for MTSF. 
Our findings indicate that re-evaluating the input time-series with a mapping function $h$ can lead to a slight improvement in model performance. 
Simultaneously, the structural exploration of NFCL-C reveals that a simple model is adequate for reevaluating historical time-series. 
We also explored ensembles of NFCL with decomposition to enhance the performance fundamentally. 
However, the performance of NFCL-D is worse than simple NFCL connection.
Moreover, NFCL demonstrates the possibility of interpreting the input-output structure. 
It shows that NFCL prevents that even if one comprehends the reasoning behind a prediction, it may become meaningless on real-world if the rationale is inaccurate.

NFCL remains imperfect. 
The learnable parameter of $h$ for independently processing all values of $w$ and $X$ is directly proportional to the size of the time-series (precisely proportional to the length $L$ of the time-series and the number of time-series $K$).
The complexity of $w$ also increases depending on not only $K$ and $L$, but also forecasting length $T$.
To enlarge the output layer weight $w$ for huge sizes of time-series can lead to misunderstandings or hinder comprehension of the forecasting reasons.

On the other hand, we point to the iterative structure of the algorithm to ensure an independent $h$ as one of the reasons for NFCL's poor computational speedup relative to its parameters. 
Aside from parallelizing the iterative algorithm, there is no solution to replace iterative algorithms yet.
In the future, our focus will be on strategies to reduce the computational complexity of NFCL. 
Additionally, we aim to transform NFCL into a simple web application, ensuring accessibility to a broad audience, including those who are not experts in artificial intelligence.


\clearpage

\bibliographystyle{ACM-Reference-Format}
\bibliography{bib}


\begin{thebibliography}{46}


\ifx \showCODEN    \undefined \def \showCODEN     #1{\unskip}     \fi
\ifx \showDOI      \undefined \def \showDOI       #1{#1}\fi
\ifx \showISBNx    \undefined \def \showISBNx     #1{\unskip}     \fi
\ifx \showISBNxiii \undefined \def \showISBNxiii  #1{\unskip}     \fi
\ifx \showISSN     \undefined \def \showISSN      #1{\unskip}     \fi
\ifx \showLCCN     \undefined \def \showLCCN      #1{\unskip}     \fi
\ifx \shownote     \undefined \def \shownote      #1{#1}          \fi
\ifx \showarticletitle \undefined \def \showarticletitle #1{#1}   \fi
\ifx \showURL      \undefined \def \showURL       {\relax}        \fi
\providecommand\bibfield[2]{#2}
\providecommand\bibinfo[2]{#2}
\providecommand\natexlab[1]{#1}
\providecommand\showeprint[2][]{arXiv:#2}

\bibitem[Agarwal et~al\mbox{.}(2021)]%
        {nam}
\bibfield{author}{\bibinfo{person}{Rishabh Agarwal}, \bibinfo{person}{Levi Melnick}, \bibinfo{person}{Nicholas Frosst}, \bibinfo{person}{Xuezhou Zhang}, \bibinfo{person}{Ben Lengerich}, \bibinfo{person}{Rich Caruana}, {and} \bibinfo{person}{Geoffrey~E Hinton}.} \bibinfo{year}{2021}\natexlab{}.
\newblock \showarticletitle{Neural additive models: Interpretable machine learning with neural nets}.
\newblock \bibinfo{journal}{\emph{Advances in Neural Information Processing Systems}}  \bibinfo{volume}{34} (\bibinfo{year}{2021}).
\newblock


\bibitem[Alzubaidi et~al\mbox{.}(2021)]%
        {ai_app_02}
\bibfield{author}{\bibinfo{person}{Laith Alzubaidi}, \bibinfo{person}{Jinglan Zhang}, \bibinfo{person}{Amjad~J. Humaidi}, \bibinfo{person}{Ayad Al-Dujaili}, \bibinfo{person}{Ye Duan}, \bibinfo{person}{Omran Al-Shamma}, \bibinfo{person}{J. Santamar{\'i}a}, \bibinfo{person}{Mohammed~A. Fadhel}, \bibinfo{person}{Muthana Al-Amidie}, {and} \bibinfo{person}{Laith Farhan}.} \bibinfo{year}{2021}\natexlab{}.
\newblock \showarticletitle{Review of deep learning: concepts, CNN architectures, challenges, applications, future directions}.
\newblock \bibinfo{journal}{\emph{Journal of Big Data}} \bibinfo{volume}{8}, \bibinfo{number}{1} (\bibinfo{date}{31 Mar} \bibinfo{year}{2021}), \bibinfo{pages}{53}.
\newblock
\showISSN{2196-1115}
\urldef\tempurl%
\url{https://doi.org/10.1186/s40537-021-00444-8}
\showDOI{\tempurl}


\bibitem[Bento et~al\mbox{.}(2021)]%
        {timeshap}
\bibfield{author}{\bibinfo{person}{Jo\~{a}o Bento}, \bibinfo{person}{Pedro Saleiro}, \bibinfo{person}{Andr\'{e}~F. Cruz}, \bibinfo{person}{M\'{a}rio~A.T. Figueiredo}, {and} \bibinfo{person}{Pedro Bizarro}.} \bibinfo{year}{2021}\natexlab{}.
\newblock \showarticletitle{TimeSHAP: Explaining Recurrent Models through Sequence Perturbations}. In \bibinfo{booktitle}{\emph{Proceedings of the 27th ACM SIGKDD Conference on Knowledge Discovery \& Data Mining}} (Virtual Event, Singapore) \emph{(\bibinfo{series}{KDD '21})}. \bibinfo{publisher}{Association for Computing Machinery}, \bibinfo{address}{New York, NY, USA}, \bibinfo{pages}{2565–2573}.
\newblock
\showISBNx{9781450383325}
\urldef\tempurl%
\url{https://doi.org/10.1145/3447548.3467166}
\showDOI{\tempurl}


\bibitem[Chen et~al\mbox{.}(2017)]%
        {smape}
\bibfield{author}{\bibinfo{person}{Chao Chen}, \bibinfo{person}{Jamie Twycross}, {and} \bibinfo{person}{Jonathan~M Garibaldi}.} \bibinfo{year}{2017}\natexlab{}.
\newblock \showarticletitle{A new accuracy measure based on bounded relative error for time series forecasting}.
\newblock \bibinfo{journal}{\emph{PLoS One}} \bibinfo{volume}{12}, \bibinfo{number}{3} (\bibinfo{date}{March} \bibinfo{year}{2017}), \bibinfo{pages}{e0174202}.
\newblock


\bibitem[Chen(2019)]%
        {baqdataset}
\bibfield{author}{\bibinfo{person}{Song Chen}.} \bibinfo{year}{2019}\natexlab{}.
\newblock \bibinfo{title}{{Beijing Multi-Site Air-Quality Data}}.
\newblock \bibinfo{howpublished}{UCI Machine Learning Repository}.
\newblock
\newblock
\shownote{{DOI}: https://doi.org/10.24432/C5RK5G}.


\bibitem[Chen et~al\mbox{.}(2023)]%
        {time-series_survey}
\bibfield{author}{\bibinfo{person}{Zonglei Chen}, \bibinfo{person}{Minbo Ma}, \bibinfo{person}{Tianrui Li}, \bibinfo{person}{Hongjun Wang}, {and} \bibinfo{person}{Chongshou Li}.} \bibinfo{year}{2023}\natexlab{}.
\newblock \showarticletitle{Long sequence time-series forecasting with deep learning: A survey}.
\newblock \bibinfo{journal}{\emph{Information Fusion}}  \bibinfo{volume}{97} (\bibinfo{year}{2023}), \bibinfo{pages}{101819}.
\newblock
\showISSN{1566-2535}
\urldef\tempurl%
\url{https://doi.org/10.1016/j.inffus.2023.101819}
\showDOI{\tempurl}


\bibitem[Chollet(2016)]%
        {depthwisecv}
\bibfield{author}{\bibinfo{person}{François Chollet}.} \bibinfo{year}{2016}\natexlab{}.
\newblock \showarticletitle{Xception: Deep Learning with Depthwise Separable Convolutions}.
\newblock \bibinfo{journal}{\emph{2017 IEEE Conference on Computer Vision and Pattern Recognition (CVPR)}} (\bibinfo{year}{2016}), \bibinfo{pages}{1800--1807}.
\newblock


\bibitem[Cooley and Tukey(1965)]%
        {fft_ct_algo}
\bibfield{author}{\bibinfo{person}{James~W. Cooley} {and} \bibinfo{person}{John~W. Tukey}.} \bibinfo{year}{1965}\natexlab{}.
\newblock \showarticletitle{An Algorithm for the Machine Calculation of Complex Fourier Series}.
\newblock \bibinfo{journal}{\emph{Math. Comp.}} \bibinfo{volume}{19}, \bibinfo{number}{90} (\bibinfo{year}{1965}), \bibinfo{pages}{297--301}.
\newblock
\showISSN{00255718, 10886842}


\bibitem[Das et~al\mbox{.}(2023)]%
        {tide}
\bibfield{author}{\bibinfo{person}{Abhimanyu Das}, \bibinfo{person}{Weihao Kong}, \bibinfo{person}{Andrew Leach}, \bibinfo{person}{Shaan~K Mathur}, \bibinfo{person}{Rajat Sen}, {and} \bibinfo{person}{Rose Yu}.} \bibinfo{year}{2023}\natexlab{}.
\newblock \showarticletitle{Long-term Forecasting with Ti{DE}: Time-series Dense Encoder}.
\newblock \bibinfo{journal}{\emph{Transactions on Machine Learning Research}} (\bibinfo{year}{2023}).
\newblock
\showISSN{2835-8856}


\bibitem[Dehghani et~al\mbox{.}(2023)]%
        {largest_vit}
\bibfield{author}{\bibinfo{person}{Mostafa Dehghani}, \bibinfo{person}{Josip Djolonga}, \bibinfo{person}{Basil Mustafa}, \bibinfo{person}{Piotr Padlewski}, \bibinfo{person}{Jonathan Heek}, \bibinfo{person}{Justin Gilmer}, \bibinfo{person}{Andreas~Peter Steiner}, \bibinfo{person}{Mathilde Caron}, \bibinfo{person}{Robert Geirhos}, \bibinfo{person}{Ibrahim Alabdulmohsin}, \bibinfo{person}{Rodolphe Jenatton}, \bibinfo{person}{Lucas Beyer}, \bibinfo{person}{Michael Tschannen}, \bibinfo{person}{Anurag Arnab}, \bibinfo{person}{Xiao Wang}, \bibinfo{person}{Carlos Riquelme~Ruiz}, \bibinfo{person}{Matthias Minderer}, \bibinfo{person}{Joan Puigcerver}, \bibinfo{person}{Utku Evci}, \bibinfo{person}{Manoj Kumar}, \bibinfo{person}{Sjoerd~Van Steenkiste}, \bibinfo{person}{Gamaleldin~Fathy Elsayed}, \bibinfo{person}{Aravindh Mahendran}, \bibinfo{person}{Fisher Yu}, \bibinfo{person}{Avital Oliver}, \bibinfo{person}{Fantine Huot}, \bibinfo{person}{Jasmijn Bastings}, \bibinfo{person}{Mark Collier}, \bibinfo{person}{Alexey~A.
  Gritsenko}, \bibinfo{person}{Vighnesh Birodkar}, \bibinfo{person}{Cristina~Nader Vasconcelos}, \bibinfo{person}{Yi Tay}, \bibinfo{person}{Thomas Mensink}, \bibinfo{person}{Alexander Kolesnikov}, \bibinfo{person}{Filip Pavetic}, \bibinfo{person}{Dustin Tran}, \bibinfo{person}{Thomas Kipf}, \bibinfo{person}{Mario Lucic}, \bibinfo{person}{Xiaohua Zhai}, \bibinfo{person}{Daniel Keysers}, \bibinfo{person}{Jeremiah~J. Harmsen}, {and} \bibinfo{person}{Neil Houlsby}.} \bibinfo{year}{2023}\natexlab{}.
\newblock \showarticletitle{Scaling Vision Transformers to 22 Billion Parameters}. In \bibinfo{booktitle}{\emph{Proceedings of the 40th International Conference on Machine Learning}} \emph{(\bibinfo{series}{Proceedings of Machine Learning Research}, Vol.~\bibinfo{volume}{202})}, \bibfield{editor}{\bibinfo{person}{Andreas Krause}, \bibinfo{person}{Emma Brunskill}, \bibinfo{person}{Kyunghyun Cho}, \bibinfo{person}{Barbara Engelhardt}, \bibinfo{person}{Sivan Sabato}, {and} \bibinfo{person}{Jonathan Scarlett}} (Eds.). \bibinfo{publisher}{PMLR}, \bibinfo{pages}{7480--7512}.
\newblock


\bibitem[Dosovitskiy et~al\mbox{.}(2021)]%
        {vit}
\bibfield{author}{\bibinfo{person}{Alexey Dosovitskiy}, \bibinfo{person}{Lucas Beyer}, \bibinfo{person}{Alexander Kolesnikov}, \bibinfo{person}{Dirk Weissenborn}, \bibinfo{person}{Xiaohua Zhai}, \bibinfo{person}{Thomas Unterthiner}, \bibinfo{person}{Mostafa Dehghani}, \bibinfo{person}{Matthias Minderer}, \bibinfo{person}{Georg Heigold}, \bibinfo{person}{Sylvain Gelly}, \bibinfo{person}{Jakob Uszkoreit}, {and} \bibinfo{person}{Neil Houlsby}.} \bibinfo{year}{2021}\natexlab{}.
\newblock \showarticletitle{An Image is Worth 16x16 Words: Transformers for Image Recognition at Scale}. In \bibinfo{booktitle}{\emph{International Conference on Learning Representations}}.
\newblock


\bibitem[Ho and Xie(1998)]%
        {ARIMA_eg}
\bibfield{author}{\bibinfo{person}{S.L. Ho} {and} \bibinfo{person}{M. Xie}.} \bibinfo{year}{1998}\natexlab{}.
\newblock \showarticletitle{The use of ARIMA models for reliability forecasting and analysis}.
\newblock \bibinfo{journal}{\emph{Computers \& Industrial Engineering}} \bibinfo{volume}{35}, \bibinfo{number}{1} (\bibinfo{year}{1998}), \bibinfo{pages}{213--216}.
\newblock
\showISSN{0360-8352}
\urldef\tempurl%
\url{https://doi.org/10.1016/S0360-8352(98)00066-7}
\showDOI{\tempurl}


\bibitem[Ke et~al\mbox{.}(2017)]%
        {eg_forecasting_passen_01}
\bibfield{author}{\bibinfo{person}{Jintao Ke}, \bibinfo{person}{Hongyu Zheng}, \bibinfo{person}{Hai Yang}, {and} \bibinfo{person}{Xiqun~(Michael) Chen}.} \bibinfo{year}{2017}\natexlab{}.
\newblock \showarticletitle{Short-term forecasting of passenger demand under on-demand ride services: A spatio-temporal deep learning approach}.
\newblock \bibinfo{journal}{\emph{Transportation Research Part C: Emerging Technologies}}  \bibinfo{volume}{85} (\bibinfo{year}{2017}), \bibinfo{pages}{591--608}.
\newblock
\showISSN{0968-090X}
\urldef\tempurl%
\url{https://doi.org/10.1016/j.trc.2017.10.016}
\showDOI{\tempurl}


\bibitem[Kim et~al\mbox{.}(2022)]%
        {PM_forecasting}
\bibfield{author}{\bibinfo{person}{Bu-Yo Kim}, \bibinfo{person}{Yun-Kyu Lim}, {and} \bibinfo{person}{Joo~Wan Cha}.} \bibinfo{year}{2022}\natexlab{}.
\newblock \showarticletitle{Short-term prediction of particulate matter (PM10 and PM2.5) in Seoul, South Korea using tree-based machine learning algorithms}.
\newblock \bibinfo{journal}{\emph{Atmospheric Pollution Research}} \bibinfo{volume}{13}, \bibinfo{number}{10} (\bibinfo{year}{2022}), \bibinfo{pages}{101547}.
\newblock
\showISSN{1309-1042}
\urldef\tempurl%
\url{https://doi.org/10.1016/j.apr.2022.101547}
\showDOI{\tempurl}


\bibitem[Lai et~al\mbox{.}(2018)]%
        {exchange_dataset}
\bibfield{author}{\bibinfo{person}{Guokun Lai}, \bibinfo{person}{Wei-Cheng Chang}, \bibinfo{person}{Yiming Yang}, {and} \bibinfo{person}{Hanxiao Liu}.} \bibinfo{year}{2018}\natexlab{}.
\newblock \showarticletitle{Modeling Long- and Short-Term Temporal Patterns with Deep Neural Networks}. In \bibinfo{booktitle}{\emph{The 41st International ACM SIGIR Conference on Research \& Development in Information Retrieval}} (Ann Arbor, MI, USA) \emph{(\bibinfo{series}{SIGIR '18})}. \bibinfo{publisher}{Association for Computing Machinery}, \bibinfo{address}{New York, NY, USA}, \bibinfo{pages}{95–104}.
\newblock
\showISBNx{9781450356572}
\urldef\tempurl%
\url{https://doi.org/10.1145/3209978.3210006}
\showDOI{\tempurl}


\bibitem[Lea et~al\mbox{.}(2016)]%
        {tcn}
\bibfield{author}{\bibinfo{person}{Colin Lea}, \bibinfo{person}{Ren{\'e} Vidal}, \bibinfo{person}{Austin Reiter}, {and} \bibinfo{person}{Gregory~D. Hager}.} \bibinfo{year}{2016}\natexlab{}.
\newblock \showarticletitle{Temporal Convolutional Networks: A Unified Approach to Action Segmentation}. In \bibinfo{booktitle}{\emph{Computer Vision -- ECCV 2016 Workshops}}, \bibfield{editor}{\bibinfo{person}{Gang Hua} {and} \bibinfo{person}{Herv{\'e} J{\'e}gou}} (Eds.). \bibinfo{publisher}{Springer International Publishing}, \bibinfo{address}{Cham}, \bibinfo{pages}{47--54}.
\newblock
\showISBNx{978-3-319-49409-8}


\bibitem[Li and Bai(2019)]%
        {so2andpm2}
\bibfield{author}{\bibinfo{person}{Ke Li} {and} \bibinfo{person}{Kaixu Bai}.} \bibinfo{year}{2019}\natexlab{}.
\newblock \showarticletitle{Spatiotemporal Associations between {PM(2.5}) and {SO(2}) as well as {NO(2}) in China from 2015 to 2018}.
\newblock \bibinfo{journal}{\emph{Int J Environ Res Public Health}} \bibinfo{volume}{16}, \bibinfo{number}{13} (\bibinfo{date}{July} \bibinfo{year}{2019}).
\newblock


\bibitem[LIU et~al\mbox{.}(2022)]%
        {scinet}
\bibfield{author}{\bibinfo{person}{Minhao LIU}, \bibinfo{person}{Ailing Zeng}, \bibinfo{person}{Muxi Chen}, \bibinfo{person}{Zhijian Xu}, \bibinfo{person}{Qiuxia LAI}, \bibinfo{person}{Lingna Ma}, {and} \bibinfo{person}{Qiang Xu}.} \bibinfo{year}{2022}\natexlab{}.
\newblock \showarticletitle{SCINet: Time Series Modeling and Forecasting with Sample Convolution and Interaction}. In \bibinfo{booktitle}{\emph{Advances in Neural Information Processing Systems}}, \bibfield{editor}{\bibinfo{person}{S.~Koyejo}, \bibinfo{person}{S.~Mohamed}, \bibinfo{person}{A.~Agarwal}, \bibinfo{person}{D.~Belgrave}, \bibinfo{person}{K.~Cho}, {and} \bibinfo{person}{A.~Oh}} (Eds.), Vol.~\bibinfo{volume}{35}. \bibinfo{publisher}{Curran Associates, Inc.}, \bibinfo{pages}{5816--5828}.
\newblock


\bibitem[Liu et~al\mbox{.}(2018)]%
        {GBDT_eg}
\bibfield{author}{\bibinfo{person}{Song Liu}, \bibinfo{person}{Yaming Cui}, \bibinfo{person}{Yaze Ma}, {and} \bibinfo{person}{Peng Liu}.} \bibinfo{year}{2018}\natexlab{}.
\newblock \showarticletitle{Short-term Load Forecasting Based on GBDT Combinatorial Optimization}. In \bibinfo{booktitle}{\emph{2018 2nd IEEE Conference on Energy Internet and Energy System Integration (EI2)}}. \bibinfo{pages}{1--5}.
\newblock
\urldef\tempurl%
\url{https://doi.org/10.1109/EI2.2018.8582108}
\showDOI{\tempurl}


\bibitem[Liu et~al\mbox{.}(2022)]%
        {pyraformer}
\bibfield{author}{\bibinfo{person}{Shizhan Liu}, \bibinfo{person}{Hang Yu}, \bibinfo{person}{Cong Liao}, \bibinfo{person}{Jianguo Li}, \bibinfo{person}{Weiyao Lin}, \bibinfo{person}{Alex~X Liu}, {and} \bibinfo{person}{Schahram Dustdar}.} \bibinfo{year}{2022}\natexlab{}.
\newblock \showarticletitle{Pyraformer: Low-Complexity Pyramidal Attention for Long-Range Time Series Modeling and Forecasting}. In \bibinfo{booktitle}{\emph{International Conference on Learning Representations}}.
\newblock


\bibitem[Liu et~al\mbox{.}(2024)]%
        {itransformer}
\bibfield{author}{\bibinfo{person}{Yong Liu}, \bibinfo{person}{Tengge Hu}, \bibinfo{person}{Haoran Zhang}, \bibinfo{person}{Haixu Wu}, \bibinfo{person}{Shiyu Wang}, \bibinfo{person}{Lintao Ma}, {and} \bibinfo{person}{Mingsheng Long}.} \bibinfo{year}{2024}\natexlab{}.
\newblock \showarticletitle{iTransformer: Inverted Transformers Are Effective for Time Series Forecasting}. In \bibinfo{booktitle}{\emph{The Twelfth International Conference on Learning Representations}}.
\newblock


\bibitem[Loshchilov and Hutter(2019)]%
        {adamw}
\bibfield{author}{\bibinfo{person}{Ilya Loshchilov} {and} \bibinfo{person}{Frank Hutter}.} \bibinfo{year}{2019}\natexlab{}.
\newblock \showarticletitle{Decoupled Weight Decay Regularization}. In \bibinfo{booktitle}{\emph{International Conference on Learning Representations}}.
\newblock


\bibitem[Lundberg and Lee(2017)]%
        {shap}
\bibfield{author}{\bibinfo{person}{Scott~M Lundberg} {and} \bibinfo{person}{Su-In Lee}.} \bibinfo{year}{2017}\natexlab{}.
\newblock \showarticletitle{A Unified Approach to Interpreting Model Predictions}.
\newblock In \bibinfo{booktitle}{\emph{Advances in Neural Information Processing Systems 30}}, \bibfield{editor}{\bibinfo{person}{I.~Guyon}, \bibinfo{person}{U.~V. Luxburg}, \bibinfo{person}{S.~Bengio}, \bibinfo{person}{H.~Wallach}, \bibinfo{person}{R.~Fergus}, \bibinfo{person}{S.~Vishwanathan}, {and} \bibinfo{person}{R.~Garnett}} (Eds.). \bibinfo{publisher}{Curran Associates, Inc.}, \bibinfo{pages}{4765--4774}.
\newblock


\bibitem[Melo et~al\mbox{.}(2022)]%
        {eg_forecasting_elec_02}
\bibfield{author}{\bibinfo{person}{João Victor~Jales Melo}, \bibinfo{person}{George Rossany~Soares Lira}, \bibinfo{person}{Edson~Guedes Costa}, \bibinfo{person}{Antonio~F. Leite~Neto}, {and} \bibinfo{person}{Iago~B. Oliveira}.} \bibinfo{year}{2022}\natexlab{}.
\newblock \showarticletitle{Short-Term Load Forecasting on Individual Consumers}.
\newblock \bibinfo{journal}{\emph{Energies}} \bibinfo{volume}{15}, \bibinfo{number}{16} (\bibinfo{year}{2022}).
\newblock
\showISSN{1996-1073}
\urldef\tempurl%
\url{https://doi.org/10.3390/en15165856}
\showDOI{\tempurl}


\bibitem[Mounir et~al\mbox{.}(2023)]%
        {eg_forecasting_elec_01}
\bibfield{author}{\bibinfo{person}{Nada Mounir}, \bibinfo{person}{Hamid Ouadi}, {and} \bibinfo{person}{Ismael Jrhilifa}.} \bibinfo{year}{2023}\natexlab{}.
\newblock \showarticletitle{Short-term electric load forecasting using an EMD-BI-LSTM approach for smart grid energy management system}.
\newblock \bibinfo{journal}{\emph{Energy and Buildings}}  \bibinfo{volume}{288} (\bibinfo{year}{2023}), \bibinfo{pages}{113022}.
\newblock
\showISSN{0378-7788}
\urldef\tempurl%
\url{https://doi.org/10.1016/j.enbuild.2023.113022}
\showDOI{\tempurl}


\bibitem[Nie et~al\mbox{.}(2023)]%
        {patchtst}
\bibfield{author}{\bibinfo{person}{Yuqi Nie}, \bibinfo{person}{Nam H.~Nguyen}, \bibinfo{person}{Phanwadee Sinthong}, {and} \bibinfo{person}{Jayant Kalagnanam}.} \bibinfo{year}{2023}\natexlab{}.
\newblock \showarticletitle{A Time Series is Worth 64 Words: Long-term Forecasting with Transformers}. In \bibinfo{booktitle}{\emph{International Conference on Learning Representations}}.
\newblock


\bibitem[Paszke et~al\mbox{.}(2019)]%
        {pytorch}
\bibfield{author}{\bibinfo{person}{Adam Paszke}, \bibinfo{person}{Sam Gross}, \bibinfo{person}{Francisco Massa}, \bibinfo{person}{Adam Lerer}, \bibinfo{person}{James Bradbury}, \bibinfo{person}{Gregory Chanan}, \bibinfo{person}{Trevor Killeen}, \bibinfo{person}{Zeming Lin}, \bibinfo{person}{Natalia Gimelshein}, \bibinfo{person}{Luca Antiga}, \bibinfo{person}{Alban Desmaison}, \bibinfo{person}{Andreas K\"{o}pf}, \bibinfo{person}{Edward Yang}, \bibinfo{person}{Zach DeVito}, \bibinfo{person}{Martin Raison}, \bibinfo{person}{Alykhan Tejani}, \bibinfo{person}{Sasank Chilamkurthy}, \bibinfo{person}{Benoit Steiner}, \bibinfo{person}{Lu Fang}, \bibinfo{person}{Junjie Bai}, {and} \bibinfo{person}{Soumith Chintala}.} \bibinfo{year}{2019}\natexlab{}.
\newblock \bibinfo{booktitle}{\emph{PyTorch: an imperative style, high-performance deep learning library}}.
\newblock \bibinfo{publisher}{Curran Associates Inc.}, \bibinfo{address}{Red Hook, NY, USA}.
\newblock


\bibitem[Porter and Heald(2019)]%
        {o2_temp}
\bibfield{author}{\bibinfo{person}{W.~C. Porter} {and} \bibinfo{person}{C.~L. Heald}.} \bibinfo{year}{2019}\natexlab{}.
\newblock \showarticletitle{The mechanisms and meteorological drivers of the summertime ozone--temperature relationship}.
\newblock \bibinfo{journal}{\emph{Atmospheric Chemistry and Physics}} \bibinfo{volume}{19}, \bibinfo{number}{21} (\bibinfo{year}{2019}), \bibinfo{pages}{13367--13381}.
\newblock
\urldef\tempurl%
\url{https://doi.org/10.5194/acp-19-13367-2019}
\showDOI{\tempurl}


\bibitem[Radenovic et~al\mbox{.}(2022)]%
        {nbm}
\bibfield{author}{\bibinfo{person}{Filip Radenovic}, \bibinfo{person}{Abhimanyu Dubey}, {and} \bibinfo{person}{Dhruv Mahajan}.} \bibinfo{year}{2022}\natexlab{}.
\newblock \showarticletitle{Neural Basis Models for Interpretability}.
\newblock \bibinfo{journal}{\emph{arXiv:2205.14120}} (\bibinfo{year}{2022}).
\newblock


\bibitem[Rojat et~al\mbox{.}(2021)]%
        {xai_for_timeseries}
\bibfield{author}{\bibinfo{person}{Thomas Rojat}, \bibinfo{person}{Rapha{\"{e}}l Puget}, \bibinfo{person}{David Filliat}, \bibinfo{person}{Javier~Del Ser}, \bibinfo{person}{Rodolphe Gelin}, {and} \bibinfo{person}{Natalia~D{\'{\i}}az Rodr{\'{\i}}guez}.} \bibinfo{year}{2021}\natexlab{}.
\newblock \showarticletitle{Explainable Artificial Intelligence {(XAI)} on TimeSeries Data: {A} Survey}.
\newblock \bibinfo{journal}{\emph{CoRR}}  \bibinfo{volume}{abs/2104.00950} (\bibinfo{year}{2021}).
\newblock
\showeprint[arXiv]{2104.00950}


\bibitem[Sarker(2021)]%
        {ai_app_01}
\bibfield{author}{\bibinfo{person}{Iqbal~H. Sarker}.} \bibinfo{year}{2021}\natexlab{}.
\newblock \showarticletitle{Deep Learning: A Comprehensive Overview on Techniques, Taxonomy, Applications and Research Directions}.
\newblock \bibinfo{journal}{\emph{SN Computer Science}} \bibinfo{volume}{2}, \bibinfo{number}{6} (\bibinfo{date}{18 Aug} \bibinfo{year}{2021}), \bibinfo{pages}{420}.
\newblock
\showISSN{2661-8907}
\urldef\tempurl%
\url{https://doi.org/10.1007/s42979-021-00815-1}
\showDOI{\tempurl}


\bibitem[Sun and You(2021)]%
        {elec_gen_contol}
\bibfield{author}{\bibinfo{person}{Li Sun} {and} \bibinfo{person}{Fengqi You}.} \bibinfo{year}{2021}\natexlab{}.
\newblock \showarticletitle{Machine Learning and Data-Driven Techniques for the Control of Smart Power Generation Systems: An Uncertainty Handling Perspective}.
\newblock \bibinfo{journal}{\emph{Engineering}} \bibinfo{volume}{7}, \bibinfo{number}{9} (\bibinfo{year}{2021}), \bibinfo{pages}{1239--1247}.
\newblock
\showISSN{2095-8099}
\urldef\tempurl%
\url{https://doi.org/10.1016/j.eng.2021.04.020}
\showDOI{\tempurl}


\bibitem[Tan and Le(2021)]%
        {effnet_v2}
\bibfield{author}{\bibinfo{person}{Mingxing Tan} {and} \bibinfo{person}{Quoc Le}.} \bibinfo{year}{2021}\natexlab{}.
\newblock \showarticletitle{EfficientNetV2: Smaller Models and Faster Training}. In \bibinfo{booktitle}{\emph{Proceedings of the 38th International Conference on Machine Learning}} \emph{(\bibinfo{series}{Proceedings of Machine Learning Research}, Vol.~\bibinfo{volume}{139})}, \bibfield{editor}{\bibinfo{person}{Marina Meila} {and} \bibinfo{person}{Tong Zhang}} (Eds.). \bibinfo{publisher}{PMLR}, \bibinfo{pages}{10096--10106}.
\newblock


\bibitem[Theissler et~al\mbox{.}(2022)]%
        {exai_time_01}
\bibfield{author}{\bibinfo{person}{Andreas Theissler}, \bibinfo{person}{Francesco Spinnato}, \bibinfo{person}{Udo Schlegel}, {and} \bibinfo{person}{Riccardo Guidotti}.} \bibinfo{year}{2022}\natexlab{}.
\newblock \showarticletitle{Explainable AI for Time Series Classification: A Review, Taxonomy and Research Directions}.
\newblock \bibinfo{journal}{\emph{IEEE Access}}  \bibinfo{volume}{10} (\bibinfo{year}{2022}), \bibinfo{pages}{100700--100724}.
\newblock
\urldef\tempurl%
\url{https://doi.org/10.1109/ACCESS.2022.3207765}
\showDOI{\tempurl}


\bibitem[Tu(1996)]%
        {adv_nn}
\bibfield{author}{\bibinfo{person}{J~V Tu}.} \bibinfo{year}{1996}\natexlab{}.
\newblock \showarticletitle{Advantages and disadvantages of using artificial neural networks versus logistic regression for predicting medical outcomes}.
\newblock \bibinfo{journal}{\emph{J Clin Epidemiol}} \bibinfo{volume}{49}, \bibinfo{number}{11} (\bibinfo{date}{Nov.} \bibinfo{year}{1996}), \bibinfo{pages}{1225--1231}.
\newblock


\bibitem[Vaswani et~al\mbox{.}(2017)]%
        {transformer}
\bibfield{author}{\bibinfo{person}{Ashish Vaswani}, \bibinfo{person}{Noam Shazeer}, \bibinfo{person}{Niki Parmar}, \bibinfo{person}{Jakob Uszkoreit}, \bibinfo{person}{Llion Jones}, \bibinfo{person}{Aidan~N. Gomez}, \bibinfo{person}{\L{}ukasz Kaiser}, {and} \bibinfo{person}{Illia Polosukhin}.} \bibinfo{year}{2017}\natexlab{}.
\newblock \showarticletitle{Attention is All You Need}. In \bibinfo{booktitle}{\emph{Proceedings of the 31st International Conference on Neural Information Processing Systems}} (Long Beach, California, USA) \emph{(\bibinfo{series}{NIPS'17})}. \bibinfo{publisher}{Curran Associates Inc.}, \bibinfo{address}{Red Hook, NY, USA}, \bibinfo{pages}{6000–6010}.
\newblock
\showISBNx{9781510860964}


\bibitem[Veerappa et~al\mbox{.}(2022)]%
        {exai_time_02}
\bibfield{author}{\bibinfo{person}{Manjunatha Veerappa}, \bibinfo{person}{Mathias Anneken}, \bibinfo{person}{Nadia Burkart}, {and} \bibinfo{person}{Marco~F. Huber}.} \bibinfo{year}{2022}\natexlab{}.
\newblock \showarticletitle{Validation of XAI explanations for multivariate time series classification in the maritime domain}.
\newblock \bibinfo{journal}{\emph{Journal of Computational Science}}  \bibinfo{volume}{58} (\bibinfo{year}{2022}), \bibinfo{pages}{101539}.
\newblock
\showISSN{1877-7503}
\urldef\tempurl%
\url{https://doi.org/10.1016/j.jocs.2021.101539}
\showDOI{\tempurl}


\bibitem[Venkatachalam et~al\mbox{.}(2023)]%
        {weather_forecasting}
\bibfield{author}{\bibinfo{person}{K. Venkatachalam}, \bibinfo{person}{Pavel Trojovský}, \bibinfo{person}{Dragan Pamucar}, \bibinfo{person}{Nebojsa Bacanin}, {and} \bibinfo{person}{Vladimir Simic}.} \bibinfo{year}{2023}\natexlab{}.
\newblock \showarticletitle{DWFH: An improved data-driven deep weather forecasting hybrid model using Transductive Long Short Term Memory (T-LSTM)}.
\newblock \bibinfo{journal}{\emph{Expert Systems with Applications}}  \bibinfo{volume}{213} (\bibinfo{year}{2023}), \bibinfo{pages}{119270}.
\newblock
\showISSN{0957-4174}
\urldef\tempurl%
\url{https://doi.org/10.1016/j.eswa.2022.119270}
\showDOI{\tempurl}


\bibitem[Wu et~al\mbox{.}(2023)]%
        {timesnet}
\bibfield{author}{\bibinfo{person}{Haixu Wu}, \bibinfo{person}{Tengge Hu}, \bibinfo{person}{Yong Liu}, \bibinfo{person}{Hang Zhou}, \bibinfo{person}{Jianmin Wang}, {and} \bibinfo{person}{Mingsheng Long}.} \bibinfo{year}{2023}\natexlab{}.
\newblock \showarticletitle{TimesNet: Temporal 2D-Variation Modeling for General Time Series Analysis}. In \bibinfo{booktitle}{\emph{International Conference on Learning Representations}}.
\newblock


\bibitem[Wu et~al\mbox{.}(2021)]%
        {autoformer}
\bibfield{author}{\bibinfo{person}{Haixu Wu}, \bibinfo{person}{Jiehui Xu}, \bibinfo{person}{Jianmin Wang}, {and} \bibinfo{person}{Mingsheng Long}.} \bibinfo{year}{2021}\natexlab{}.
\newblock \showarticletitle{Autoformer: Decomposition Transformers with {Auto-Correlation} for Long-Term Series Forecasting}. In \bibinfo{booktitle}{\emph{Advances in Neural Information Processing Systems}}.
\newblock


\bibitem[Xu et~al\mbox{.}(2019)]%
        {layernorm}
\bibfield{author}{\bibinfo{person}{Jingjing Xu}, \bibinfo{person}{Xu Sun}, \bibinfo{person}{Zhiyuan Zhang}, \bibinfo{person}{Guangxiang Zhao}, {and} \bibinfo{person}{Junyang Lin}.} \bibinfo{year}{2019}\natexlab{}.
\newblock \showarticletitle{Understanding and Improving Layer Normalization}. In \bibinfo{booktitle}{\emph{Advances in Neural Information Processing Systems}}, \bibfield{editor}{\bibinfo{person}{H.~Wallach}, \bibinfo{person}{H.~Larochelle}, \bibinfo{person}{A.~Beygelzimer}, \bibinfo{person}{F.~d\textquotesingle Alch\'{e}-Buc}, \bibinfo{person}{E.~Fox}, {and} \bibinfo{person}{R.~Garnett}} (Eds.), Vol.~\bibinfo{volume}{32}. \bibinfo{publisher}{Curran Associates, Inc.}
\newblock


\bibitem[Yoshikado(2023)]%
        {wind_o2_temp}
\bibfield{author}{\bibinfo{person}{Hiroshi Yoshikado}.} \bibinfo{year}{2023}\natexlab{}.
\newblock \showarticletitle{Correlation between air temperature and surface ozone in their extreme ranges in the greater Tokyo region}.
\newblock \bibinfo{journal}{\emph{Asian Journal of Atmospheric Environment}} \bibinfo{volume}{17}, \bibinfo{number}{1} (\bibinfo{date}{Aug.} \bibinfo{year}{2023}), \bibinfo{pages}{9}.
\newblock


\bibitem[Zeng et~al\mbox{.}(2023)]%
        {dnlinear}
\bibfield{author}{\bibinfo{person}{Ailing Zeng}, \bibinfo{person}{Muxi Chen}, \bibinfo{person}{Lei Zhang}, {and} \bibinfo{person}{Qiang Xu}.} \bibinfo{year}{2023}\natexlab{}.
\newblock \showarticletitle{Are Transformers Effective for Time Series Forecasting?}
\newblock \bibinfo{journal}{\emph{Proceedings of the AAAI Conference on Artificial Intelligence}}.
\newblock


\bibitem[Zhang et~al\mbox{.}(2024)]%
        {eg_forecasting_wind_01}
\bibfield{author}{\bibinfo{person}{Chu Zhang}, \bibinfo{person}{Yuhan Wang}, \bibinfo{person}{Yongyan Fu}, \bibinfo{person}{Xiujie Qiao}, \bibinfo{person}{Muhammad~Shahzad Nazir}, {and} \bibinfo{person}{Tian Peng}.} \bibinfo{year}{2024}\natexlab{}.
\newblock \showarticletitle{A novel DWTimesNet-based short-term multi-step wind power forecasting model using feature selection and auto-tuning methods}.
\newblock \bibinfo{journal}{\emph{Energy Conversion and Management}}  \bibinfo{volume}{301} (\bibinfo{year}{2024}), \bibinfo{pages}{118045}.
\newblock
\showISSN{0196-8904}
\urldef\tempurl%
\url{https://doi.org/10.1016/j.enconman.2023.118045}
\showDOI{\tempurl}


\bibitem[Zhou et~al\mbox{.}(2020)]%
        {informer}
\bibfield{author}{\bibinfo{person}{Haoyi Zhou}, \bibinfo{person}{Shanghang Zhang}, \bibinfo{person}{Jieqi Peng}, \bibinfo{person}{Shuai Zhang}, \bibinfo{person}{Jianxin Li}, \bibinfo{person}{Hui Xiong}, {and} \bibinfo{person}{Wan Zhang}.} \bibinfo{year}{2020}\natexlab{}.
\newblock \showarticletitle{Informer: Beyond Efficient Transformer for Long Sequence Time-Series Forecasting}. In \bibinfo{booktitle}{\emph{AAAI Conference on Artificial Intelligence}}.
\newblock


\bibitem[Zhou et~al\mbox{.}(2022)]%
        {fedformer}
\bibfield{author}{\bibinfo{person}{Tian Zhou}, \bibinfo{person}{Ziqing Ma}, \bibinfo{person}{Qingsong Wen}, \bibinfo{person}{Xue Wang}, \bibinfo{person}{Liang Sun}, {and} \bibinfo{person}{Rong Jin}.} \bibinfo{year}{2022}\natexlab{}.
\newblock \showarticletitle{{FEDformer}: Frequency enhanced decomposed transformer for long-term series forecasting}. In \bibinfo{booktitle}{\emph{Proc. 39th International Conference on Machine Learning (ICML 2022)}} (Baltimore, Maryland).
\newblock


\end{thebibliography}



\clearpage

\appendix


\section*{Appendix}

\section{NFCL-C implementation}\label{app:nfcl-implementation}

\begin{algorithm}[ht]
    \caption{Iterative re-evaluation for complex NFCL}\label{alg:iter_nfcl_c}
\begin{algorithmic}

    \Require $X \in \mathbb{R}^{K, L}$, $H=[[h^{1,1}, h^{1,2}, \cdots h^{1,L}]\times K]$
    \Ensure $X_h \in \mathbb{R}^{K, L}$
    
    \State $X_h \gets $ copy$(X)$.zero()\Comment{the zero array has same shape as $X$.}
    \For{$k$ \textbf{in} $1,2,\cdots, K$}
        \For{$l$ \textbf{in} $1,2,\cdots, L$}
            \State $h^{k,l} \gets H[k][l]$\Comment{$h^{k,l}$ is the non-linear mapping function.}
            \State $X_h[k][l] \gets h^{k,l}(X[k][l])$
        
        \EndFor
    \EndFor
\end{algorithmic}
\end{algorithm}

\begin{algorithm}[ht]
    \caption{Group convolution for complex NFCL}\label{alg:group_conv_nfcl_c}
\begin{algorithmic}
    \Require $X \in \mathbb{R}^{K, L}$, $H=[[h^{1,1}, h^{1,2}, \cdots h^{1,L}]\times K]$
    \Require $C = [c_1, \cdots, c_N]$, $\gamma = K\cdot L$ \Comment{argument $\gamma$ is group size.}
    \Ensure $X_h \in \mathbb{R}^{K, L}$

    \State $X_h \gets \textrm{Flatten}(X) \in \mathbb{R}^{K \cdot L, 1}$
    \State $C_{in} \gets [1] + [c_1, \cdots, c_{N-1}]$ ; $C_{out} \gets [c_1, \cdots, c_N]$
    \For{$c_{in}, c_{out} \ \textbf{in} \ C_{in}, C_{out}$}
        \State $X_{h} \gets Conv(X_h, c_{i}\cdot \gamma, c_{o}\cdot \gamma, \gamma )$
        \State $X_h \gets \mathrm{activation}(X_h)$
    \EndFor
    \State $X_{h} \gets Conv(X_h, c_N\cdot \gamma, 1\cdot \gamma, \gamma)$ 
\end{algorithmic}
\end{algorithm}

NFCL-V computes predictions for all future time-series points by multiplying weights from all past time-series points. 
NFCL-C, on the other hand, involves a simple process: before multiplying the weights of all past time-series points, each individual time-series point undergoes re-evaluation by a trainable NNs. 
To uphold our proposed concept, it's crucial that the NN responsible for each time-series point remains unaffected by other points. 
Consequently, the Equation \ref{eq:mapping_ours} is satisfied by the below Algorithm \ref{alg:iter_nfcl_c} for NFCL-C.
In the Algorithm \ref{alg:iter_nfcl_c}, $H$ denotes the $K\cdot L$ number of mapping function $h$, and $x_h$ is re-evaluated value by each $h$.
As a result of the Algorithm, we obtain $X_h$ a re-evaluation of the input, which can substitute $X$ in Equation \ref{eq:simple_ours}. 
If we substitute $h^{i,j}\left(x^{i,j}\right)$ in Equation \ref{eq:mapping_ours} with $X_h$, we observe that Equations \ref{eq:simple_ours} and \ref{eq:mapping_ours} are fully compatible.

To enhance the efficiency of the Algorithm during double iterations, we hired the group convolution. 
Widely employed in image processing, the group convolution partitions the input channel into multiple groups, establishing independent CNN to execute convolutional operations exclusively on these groups. 
Specifically, when the size of the image region matches the number of groups, a CNN is independently created for each pixel in every image to carry out the operation, and this is called the point-wise convolution. 
We leverage this approach to construct a re-evaluation module $H$ that independently operates at each time point for every time-series variable.
In the Algorithm \ref{alg:group_conv_nfcl_c}, the parameter of the function $Conv(X_h, c_{i}\cdot \gamma, c_{o}\cdot \gamma, \gamma)$ sequentially denotes the input tensor, the number of input tensor's channel, the number of output tensor's channels, and the group size, respectively. 
Since we set the same group size $\gamma$ to the number of the first input tensor's channel, we can easily implement the point-wise convolution.

\section{Decomposition algorithm}\label{app:decposition_algorithm}


\begin{algorithm}[ht]
    \caption{Decomposition for the time-series}\label{alg:decomp_ts}
\begin{algorithmic}[0]

    \Require $X \in \mathbb{R}^{K,L}, S = {s_1, \cdots, s_N}$
    \Ensure $X_{dec} \in \mathbb{R}^{K, L}$

    \State $X_{dec} \gets []$
    \While{$S\neq \varnothing $}
    
        \State $s_n \gets S\mathrm{.pop()}$
        \State $X_{pad} \gets \mathrm{LeftPad}(X, s_n - 1)$
        \State $X_{s_n} \gets \mathrm{AveragePool}(X_{pad}, s_n)$
        \State $X_{dec}\mathrm{.append}(X_{s_n})$
        \State $X \gets X - X_{s_n}$
        
    \EndWhile
\end{algorithmic}
\end{algorithm}

\begin{algorithm}[ht]
    \caption{Workflow of NFCL-D}\label{alg:nfcl-d}
\begin{algorithmic}[0]

    \Require $X_{dec} \in \mathbb{R}^{K, L}$, $M = [\mathrm{NFCL-C}_{s_1}, \cdots, \mathrm{NFCL-C}_{s_N}]$
    \Ensure $\tilde{Y} \in \mathbb{R}^{K, T}$

    \State $\hat{Y}_{dec} \gets []$
    \While{$M \neq \varnothing \mathrm{\ and\ } X_{dec} \neq \varnothing$}

        \State $m_{s_n} \gets M\mathrm{.pop()}$
        \State $X_{s_n} \in \mathbb{R}^{K,L} \gets X_{dec}\mathrm{.pop()}$
        \State $\hat{Y}_{s_n} \gets m_{s_n}\left( X_{s_n} \right)$
        \State $\hat{Y}_{dec}\mathrm{.append}(\hat{Y}_{s_n})$
        
    \EndWhile

    \State $\tilde{Y} \gets \mathrm{Sum}\left(\mathrm{Concat}\left(\hat{Y}_{dec}\right)\right)$
    
\end{algorithmic}
\end{algorithm}

We describe the process of NFCL-D. 
Algorithms \ref{alg:decomp_ts} and \ref{alg:nfcl-d} provide the sequences of the pipeline.
Algorithm \ref{alg:decomp_ts} depicts the decomposition process.
And Algorithm \ref{alg:nfcl-d} displays how to forecast the future using NFCL-D.
Since we employed $\left|S\right|$ times NFCL-Cs, the number of NFCL-D's parameters depends on $\left|S\right|$.

In the Algorithm \ref{alg:decomp_ts}, the LeftPad function copies the first time value in the time-series and pads it to the left of the given $X$ by the length specified as the second argument $s_n-1$.
Similarly, the AveragePool function rolls and averages the $X$ using the kernel size $s_n$.
In the Algorithm \ref{alg:nfcl-d}, the Sum and Concat functions perform as the Python general function way, such as NumPy.
Two functions focused to sum between the $\hat{Y}_{dec}$ that used different kernel sizes.
Consequently, we can forecast the $\hat{Y}$ to recover the scale of $\tilde{Y}$.

\begin{table*}[t]
\centering
\begin{tabular}{cccccccc} \toprule \toprule
Category                     & Dataset          & Abbreviation & Freq.                    & \# of var.          & \# of Train               & \# of Val.               & \# of Test.               \\ \midrule \midrule
Patient                      & National illness & ILL          & 7 days                   & 7                   & 532                    & 171                    & 170                    \\ \midrule
Currency exchange            & Exchange   & Exchange     & 1 day                    & 8                   & 4505                   & 1496                   & 1494                   \\ \midrule
\multirow{4}{*}{Electricity} & ETTh1            & ETTh1        & \multirow{2}{*}{1 hour}  & \multirow{2}{*}{7}  & \multirow{2}{*}{10405} & \multirow{2}{*}{3461}  & \multirow{2}{*}{3461}  \\
                             & ETTh2            & ETTh2        &                          &                     &                        &                        &                        \\ \cmidrule{2-8}
                             & ETTm1            & ETTm1        & \multirow{2}{*}{15 min.} & \multirow{2}{*}{7}  & \multirow{2}{*}{41761} & \multirow{2}{*}{13913} & \multirow{2}{*}{13913} \\ 
                             & ETTm2            & ETTm2        &                          &                     &                        &                        &                        \\ \midrule
\multirow{9}{*}{Climate}     & Aotizhongxin     & Aotiz.       & \multirow{8}{*}{1 hour}  & \multirow{8}{*}{10} & \multirow{8}{*}{20991} & \multirow{8}{*}{6991}  & \multirow{8}{*}{6989}  \\
                             & Changping        & Chang.       &                          &                     &                        &                        &                        \\
                             & Dingling         & Ding.        &                          &                     &                        &                        &                        \\
                             & Dongsi           & Dongsi       &                          &                     &                        &                        &                        \\
                             & Guanyuan         & Guan.        &                          &                     &                        &                        &                        \\
                             & Gucheng          & Guch.        &                          &                     &                        &                        &                        \\
                             & Huairou          & Huai.        &                          &                     &                        &                        &                        \\
                             & Nongzhanguan     & Nong.        &                          &                     &                        &                        &                        \\ \cmidrule{2-8}
                             & Weather          & Weather      & 10 min.                  & 21                  & 31570                  & 10517                  & 10516                 \\ \bottomrule \bottomrule
\end{tabular}%
\caption{Details of datasets.}
\label{tab:detail_data}
\end{table*}

\section{Detail of datasets}\label{app:detail_dataset}


We used the online-available open datasets from the UCI-repository, and GitHubs. 
Autoformer\footnote{https://github.com/thuml/Autoformer/tree/main} aligned some datasets on their repo: Weather, National illness, and Currency Exchange.
In particular, three datasets that Autoformer provided applied underwent minimal preprocessing.
We got the others, electricity transfer temperature and Beijing air quality, from its GitHub\footnote{https://github.com/zhouhaoyi/ETDataset/tree/main/ETT-small} or UCI-repository\footnote{https://archive.ics.uci.edu/dataset/501/beijing+multi+site+air+quality+data}. 
The details of datasets are Table \ref{tab:detail_data}, and briefly summaries are as follow:

\begin{itemize}
    \item Weather\footnote{https://www.bgc-jena.mpg.de/wetter/} dataset has 21 climate records, such as temperature, rainfall, and etc. This dataset had collected every 10 minute on 2020 year.
    \item National illness\footnote{https://gis.cdc.gov/grasp/fluview/fluportaldashboard.html} dataset was collected by Centers for Disease Control and Prevention of the United States between 2002 to 2020. This dataset targeted to the influenza-like illness patients. Note that we use the abbreviation as \bd{ILL}.
    \item Exchange dataset records the daily currency exchange rates of eight nations (Australia, British, Canada, Switzerland, China, Japan, NewZealand, and Singapore) from 1990 to 2010 \cite{exchange_dataset}.
    \item Note that we use the abbreviation of the electricity transfer temperature as \bd{ETT}. ETT recorded the oil temperature and the electricity's loads, it was recorded every 15 minutes or hourly between 2016 to 2018. Also, ETT has four sub-dataset, becuase ETT was collected from two different provinces and has different acquisition frequency. A number of ETT indicates the province as "1" or "2". Thus, "m" and "h" denotes the frequency of ETT collected.
    \item Beijing air quality is the record of subdistrict's climates between 2013--2017. In contrast to the weather dataset, it also forcused on the two different sizes' inhalable particulate matters. In preprocessing, we drop the time-series variables; the wind direction and rainfall.

\end{itemize}

\section{Metrics}\label{app:metrics}

We held the manner of Informer experiments for MTSF \cite{informer}.
There are two manners; one is the sampling the dataset with the one step sliding and the other is the method how to evaluate the MTSF performance. The evaluation process performed on zero-mean scaled time-series variables.
Kept the notation, we compute the each metrics as,

\begin{enumerate}
    \item MAE
        $$ MAE(Y', \hat{Y}') = \frac{1}{|D'| \cdot K} \sum_{n=1}^{|D'|} \sum_{k=1}^{K} \left| y_n^{k,:} - \hat{y}_n^{k,:}  \right|$$
       \item MSE
        $$ MAE(Y', \hat{Y}') = \frac{1}{|D'| \cdot K} \sum_{n=1}^{|D'|} \sum_{k=1}^{K} \left( y_n^{k,:} - \hat{y}_n^{k,:}  \right)^2$$
    \item SMAPE
        $$ SMAPE(Y', \hat{Y}') = \frac{100\%}{|D'| \cdot K} \sum_{n=1}^{|D'|} \sum_{k=1}^{K} \frac{\left|  y_n^{k,:} - \hat{y}_n^{k,:} \right|}{ \left|  y_n^{k,:} \right| + \left|  \hat{y}_n^{k,:} \right| }$$
    \item $R^2$
        $$R^2(Y', \hat{Y}') = 1 - \frac{\sum_{n=1}^{|D'|}{\sum_{k=1}^{K}{(y_n^{k,:} - \hat{y}_n^{k,:})^2}}}{\sum_{n=1}^{|D'|}{\sum_{k=1}^{K}{(y_n^{k,:} - \bar{y}^{k,:})^2}}}$$
\end{enumerate}
where, $\bar{Y}'$ and $|D'|$ denote the average of test $Y'$ and the number of samples in $|D'|$.


\section{hyperparameter}\label{app:hyper_params}

Basically, we applied to the baseline's hyperparameter based on the PatchTST experiments\footnote{https://github.com/yuqinie98/PatchTST/tree/main/PatchTST\_supervised/models}.
Since their Github provides seven baselines (DLinear, NLinear, Informer, Autoformer, PatchTST, Linear, and Transformer), we conducted the experiments to use their baselines.
However, the Transformer and simple-Linear layer showed too much degraded performance, so we excluded the summary of the scores.
Moreover, these baselines didn't employ the CNN-based model, we referred to the official GitHub of SCINet\footnote{https://github.com/cure-lab/SCINet}, TimesNet\footnote{https://github.com/thuml/Time-Series-Library}, and iTransformer\footnote{https://github.com/thuml/iTransformer}.
We also tried to employ the MICN, and TiDE from the GitHub that also shared the TimesNet baseline.
While the TiDE shows competitive performance, MICN implementation didn't work.
So, we included only the TiDE in our comparisons.
In addition, we double-check the hyperparameter setting from their papers.
The table \ref{tab:hp_setting} records the full setting.
And, a brief abstraction of each baseline's properties is provided below:

\begin{itemize}
    \item \bd{LSTF-Linear} \cite{dnlinear} proposed two baselines; where letters D and N denote decomposition and normalization, respectively, with fully-connected weights for univariate time-series.
    \item \bd{TiDE} \cite{tide} introduced an encoder-decoder forecasting approach based on linear combination stacking.
    \item \bd{Informer} \cite{informer} serves as a fundamental method utilizing the transformer architecture. Informer focuses on reducing computational costs and provides an open dataset.
    \item \bd{Autoformer} \cite{autoformer} is composed of three attention blocks for encoding, trend decoding, and seasonal decoding. To effectively capture auto-correlation, it employs Fourier transformation and aggregation modules in each block.
    \item \bd{PatchTST} \cite{patchtst} is a transformer-based method forecasting channels independently. Additionally, it incorporates a patching process to enhance forecasting performance.
    \item \bd{SCINet} \cite{scinet} remains a state-of-the-art model in electricity temperature transfer datasets \cite{informer}. It forecasts the future based on sample convolution and interaction blocks that function similarly to the Fourier transform \cite{fft_ct_algo}.
    \item \bd{TimesNet} \cite{timesnet} utilizes CNN and the Fourier transform to extract information from the frequency-amplitude map.
    \item \bd{iTransformer} \cite{itransformer} simply applies the attention and feed-forward network on the transposed dimensions of time and variables. The attention mechanism captured multivariate correlations instead of the time-step correlations. 
\end{itemize}

\begin{table}[t]
\centering
\resizebox{\columnwidth}{!}{%
\begin{tabular}{llllll} \toprule \toprule
Model                     & Params.      & Setting & Model                      & Params.        & Setting \\ \midrule \midrule
\multirow{4}{*}{TimesNet} & top\_k       & 5       & \multirow{4}{*}{SCINet}    & hiddens        & 1       \\
                          & kernel\_size & 6       &                            & kernel\_size   & 5       \\
                          & d\_ff        & 16      &                            & dropout        & 0.5     \\
                          & d\_ff        & 64      &                            & group          & 1       \\ \midrule 
\multirow{11}{*}{Former}  & moving\_avg  & 25      & \multirow{11}{*}{PatchTST} & fc\_dropout    & 0.05    \\
                          & factor       & 1       &                            & head\_dropout  & 0       \\
                          & dropout      & 0.05    &                            & patch\_len     & 16      \\
                          & embed        & timeF   &                            & stride         & 8       \\
                          & activation   & gelu    &                            & padding\_patch & end     \\
                          & embed\_type  & 0       &                            & subtract\_last & 0       \\
                          & d\_model     & 512     &                            & decomposition  & 0       \\
                          & n\_heads     & 8       &                            & kernel\_size   & 25      \\
                          & e\_layers    & 2       &                            & individual     & 0       \\
                          & d\_layers    & 1       &                            &                &         \\
                          & d\_ff        & 2048    &                            &                &         \\ \bottomrule \bottomrule
\end{tabular}%
}
\caption{hyperparameter setting.}
\label{tab:hp_setting}
\end{table}

\section{NFCL Inter-comparison}\label{baseline-comparison}

We only included the rankings between nine benchmarks and NFCL-C with $h=32$ in the main content of this section.
To present our complete comparisons, we have added an additional section with long tables.
The title of each table indicates the evaluation metric. 
These tables are summarized in Table \ref{tab:summary_benchmark} in Section \ref{sec:compare_all}.

\begin{table*}[ht]
\centering
\resizebox{.75\textwidth}{!}{%
\begin{tabular}{cccccccccccc} \toprule \toprule
Dataset                    & $T$        & NFCL                                  & DLinear                               & NLinear     & TiDE                                  & Informer & Autoformer & PatchTST                              & iTransformer                          & SCINet                                & TimesNet    \\ \cmidrule(lr){1-2} \cmidrule(lr){3-12} \morecmidrules \cmidrule(lr){1-2} \cmidrule(lr){3-12}
                           & 6          & {\color[HTML]{0000FF} \textbf{0.070}} & 0.099                                 & 0.080       & 0.084                                 & 0.187    & 0.146      & \ul{0.072}                           & 0.105                                 & 0.089                                 & 0.082       \\
                           & 12         & {\color[HTML]{0000FF} \textbf{0.090}} & 0.131                                 & 0.106       & 0.108                                 & 0.236    & 0.175      & \ul{0.096}                           & 0.128                                 & 0.115                                 & 0.102       \\
                           & 18         & {\color[HTML]{0000FF} \textbf{0.108}} & 0.160                                 & 0.129       & 0.128                                 & 0.296    & 0.200      & \ul{0.116}                           & 0.148                                 & 0.143                                 & 0.126       \\
\multirow{-4}{*}{Weather}  & 24         & {\color[HTML]{0000FF} \textbf{0.125}} & 0.186                                 & 0.148       & 0.146                                 & 0.412    & 0.223      & \ul{0.136}                           & 0.162                                 & 0.169                                 & 0.146       \\
\bottomrule \bottomrule
\end{tabular}%
}
\caption{MAE ($\downarrow$) comparison among NFCL and nine benchmarks.}
\label{tab:MAE-bench-tb}
\end{table*}

\begin{table*}[]
\centering
\resizebox{.75\textwidth}{!}{%
\begin{tabular}{cccccccccccc} \toprule \toprule
Dataset                    & $T$        & NFCL                                  & DLinear                               & NLinear     & TiDE        & Informer    & Autoformer  & PatchTST                              & iTransformer                          & SCINet                                & TimesNet    \\ \cmidrule(lr){1-2} \cmidrule(lr){3-12} \morecmidrules \cmidrule(lr){1-2} \cmidrule(lr){3-12}
                           & 6          & \ul{0.058}                           & 0.064                                 & 0.063       & 0.063       & 0.111       & 0.090       & 0.060                                 & 0.075                                 & {\color[HTML]{0000FF} \textbf{0.056}} & 0.066       \\
                           & 12         & \ul{0.075}                           & 0.087                                 & 0.087       & 0.085       & 0.174       & 0.115       & 0.080                                 & 0.098                                 & {\color[HTML]{0000FF} \textbf{0.074}} & 0.082       \\
                           & 18         & {\color[HTML]{0000FF} \textbf{0.089}} & 0.108                                 & 0.106       & 0.105       & 0.248       & 0.135       & 0.097                                 & 0.117                                 & \ul{0.092}                           & 0.106       \\
\multirow{-4}{*}{Weather}  & 24         & {\color[HTML]{0000FF} \textbf{0.102}} & 0.127                                 & 0.123       & 0.121       & 0.458       & 0.155       & 0.113                                 & 0.131                                 & \ul{0.110}                           & 0.122       \\
\bottomrule \bottomrule     
\end{tabular}%
}
\caption{MSE ($\downarrow$) comparison among NFCL and nine benchmarks.}
\label{tab:MSE-bench-tb}
\end{table*}

\begin{table*}[]
\centering
\resizebox{.75\textwidth}{!}{%
\begin{tabular}{cccccccccccc} \toprule \toprule
Dataset                    & $T$        & NFCL                                   & DLinear      & NLinear                                & TiDE                                  & Informer & Autoformer & PatchTST                               & iTransformer                           & SCINet       & TimesNet                               \\ \cmidrule(lr){1-2} \cmidrule(lr){3-12} \morecmidrules \cmidrule(lr){1-2} \cmidrule(lr){3-12}
                           & 6          & \ul{12.280}                           & 15.050       & 13.406                                 & 13.857                                & 29.537   & 25.124     & {\color[HTML]{0000FF} \textbf{12.264}} & 16.422                                 & 17.493       & 13.970                                 \\
                           & 12         & {\color[HTML]{0000FF} \textbf{14.712}} & 18.482       & 16.264                                 & 16.538                                & 33.287   & 28.120     & \ul{15.165}                           & 18.863                                 & 21.764       & 16.322                                 \\
                           & 18         & {\color[HTML]{0000FF} \textbf{16.706}} & 21.657       & 18.778                                 & 18.619                                & 38.851   & 31.693     & \ul{17.373}                           & 20.830                                 & 25.141       & 18.581                                 \\
\multirow{-4}{*}{Weather}  & 24         & {\color[HTML]{0000FF} \textbf{18.467}} & 24.498       & 20.573                                 & 20.329                                & 45.846   & 33.693     & \ul{19.458}                           & 22.083                                 & 28.143       & 20.469                                 \\
\bottomrule \bottomrule
\end{tabular}%
}
\caption{SMAPE ($\downarrow$) comparison among NFCL and nine benchmarks.}
\label{tab:SMAPE-bench-tb}
\end{table*}

\begin{table*}[]
\centering
\resizebox{.75\textwidth}{!}{%
\begin{tabular}{cccccccccccc} \toprule \toprule
Dataset                    & $T$        & NFCL                                  & DLinear & NLinear                               & TiDE        & Informer & Autoformer                            & PatchTST                              & iTransformer                          & SCINet                                & TimesNet    \\ \cmidrule(lr){1-2} \cmidrule(lr){3-12} \morecmidrules \cmidrule(lr){1-2} \cmidrule(lr){3-12}
                           & 6          & \ul{0.822}                           & 0.807   & 0.810                                 & 0.809       & -0.172   & -0.059                                & {\color[HTML]{0000FF} \textbf{0.826}} & 0.763                                 & 0.763                                 & 0.797       \\
                           & 12         & {\color[HTML]{0000FF} \textbf{0.772}} & 0.729   & 0.730                                 & 0.732       & -0.656   & -0.068                                & \ul{0.756}                           & 0.684                                 & 0.680                                 & 0.735       \\
                           & 18         & {\color[HTML]{0000FF} \textbf{0.717}} & 0.652   & 0.655                                 & 0.661       & -1.109   & -0.074                                & \ul{0.688}                           & 0.610                                 & 0.577                                 & 0.645       \\
\multirow{-4}{*}{Weather}  & 24         & {\color[HTML]{0000FF} \textbf{0.659}} & 0.580   & 0.585                                 & 0.595       & -2.933   & -0.316                                & \ul{0.619}                           & 0.555                                 & 0.453                                 & 0.560       \\
\bottomrule \bottomrule
\end{tabular}%
}
\caption{$R^2$ ($\uparrow$) comparison among NFCL and nine benchmarks.}
\label{tab:r2-bench-tb}
\end{table*}

\section{NFCL Intra-comparison}\label{app:nfcm_compare}

We only included the rankings of NFCL variations in the main content of this section.
To present our complete comparisons, we have added an additional section with long tables.
The title of each table indicates the evaluation metric. 
These tables are summarized in Table \ref{tab:hyper_comparison} in Section \ref{sec:compare_hyper}.

\begin{table*}[!ht]
\centering
\resizebox{.85\textwidth}{!}{%
\begin{tabular}{cccccccccccccccc} \toprule \toprule
\multicolumn{2}{c}{Variation}           & NFCL-V                                & \multicolumn{9}{c}{NFCL-C}                                                                                                                                                                                                                                                                                                                  & \multicolumn{4}{c}{NFCL-D}                                                                      \\ \cmidrule(lr){1-2} \cmidrule(lr){3-3} \cmidrule(lr){4-12} \cmidrule(lr){13-16}
Datasets                   & $T$        & -                                     & 16                                    & 32                                    & 64                                    & 128                                   & 16,16                                 & 32,32       & 64,64                                 & 128,128                               & 128,128,64                            & 64                                    & 128                                   & 64,64 & 128,128 \\ \cmidrule(lr){1-2} \cmidrule(lr){3-3} \cmidrule(lr){4-12} \cmidrule(lr){13-16} \morecmidrules \cmidrule(lr){1-2} \cmidrule(lr){3-3} \cmidrule(lr){4-12} \cmidrule(lr){13-16}
                           & 6          & 0.071                                 & 0.070                                 & 0.070                                 & 0.070                                 & {\color[HTML]{0000FF} \textbf{0.070}} & 0.070                                 & 0.070       & 0.071                                 & \ul{0.070}                           & 0.070                                 & 0.071                                 & 0.071                                 & 0.071 & 0.072   \\
\bottomrule \bottomrule
\end{tabular}%
}
\caption{MAE ($\downarrow$) comparison among hyperparameter tuning.}
\label{tab:hp_search_mae}
\end{table*}

\begin{table*}[ht]
\centering
\resizebox{.85\textwidth}{!}{%
\begin{tabular}{cccccccccccccccc} \toprule \toprule
\multicolumn{2}{c}{Variation}           & NFCL-V                                & \multicolumn{9}{c}{NFCL-C}                                                                                                                                                                                                                                                                                                                                            & \multicolumn{4}{c}{NFCL-D}                       \\ \cmidrule(lr){1-2} \cmidrule(lr){3-3} \cmidrule(lr){4-12} \cmidrule(lr){13-16}
Datasets                   & $T$        & -                                     & 16                                    & 32                                    & 64                                    & 128                                   & 16,16                                 & 32,32                                 & 64,64                                 & 128,128                               & 128,128,64                            & 64          & 128    & 64,64       & 128,128     \\ \cmidrule(lr){1-2} \cmidrule(lr){3-3} \cmidrule(lr){4-12} \cmidrule(lr){13-16} \morecmidrules \cmidrule(lr){1-2} \cmidrule(lr){3-3} \cmidrule(lr){4-12} \cmidrule(lr){13-16}
                           & 6          & \ul{0.057}                           & 0.058                                 & 0.058                                 & 0.058                                 & {\color[HTML]{0000FF} \textbf{0.057}} & 0.058                                 & 0.058                                 & 0.058                                 & 0.057                                 & 0.057                                 & 0.058       & 0.058  & 0.059       & 0.058       \\
\bottomrule \bottomrule
\end{tabular}%
}
\caption{MSE ($\downarrow$) comparison among hyperparameter tuning.}
\label{tab:hp_search_mse}
\end{table*}

\begin{table*}[ht]
\centering
\resizebox{.85\textwidth}{!}{%
\begin{tabular}{cccccccccccccccc} \toprule \toprule
\multicolumn{2}{c}{Variation}           & NFCL-V                                 & \multicolumn{9}{c}{NFCL-C}                                                                                                                                                                                                                                                                                                                         & \multicolumn{4}{c}{NFCL-D}                                               \\ \cmidrule(lr){1-2} \cmidrule(lr){3-3} \cmidrule(lr){4-12} \cmidrule(lr){13-16}
Datasets                   & $T$        & -                                      & 16                                     & 32                                     & 64                                     & 128                                    & 16,16                                  & 32,32                                  & 64,64                                  & 128,128                                & 128,128,64 & 64           & 128                                    & 64,64  & 128,128 \\ \cmidrule(lr){1-2} \cmidrule(lr){3-3} \cmidrule(lr){4-12} \cmidrule(lr){13-16} \morecmidrules \cmidrule(lr){1-2} \cmidrule(lr){3-3} \cmidrule(lr){4-12} \cmidrule(lr){13-16}
                           & 6          & 12.382                                 & 12.304                                 & {\color[HTML]{0000FF} \textbf{12.280}} & 12.299                                 & \ul{12.294}                           & 12.316                                 & 12.298                                 & 12.429                                 & 12.336                                 & 12.475     & 12.485       & 12.518                                 & 12.562 & 12.629  \\
\bottomrule \bottomrule
\end{tabular}%
}
\caption{SMAPE ($\downarrow$) comparison among hyperparameter tuning.}
\label{tab:hp_search_smape}
\end{table*}

\begin{table*}[ht]
\centering
\resizebox{.85\textwidth}{!}{%
\begin{tabular}{cccccccccccccccc} \toprule \toprule
\multicolumn{2}{c}{Variation}           & NFCL-V                                & \multicolumn{9}{c}{NFCL-C}                                                                                                                                                                                                                                                                                                                                             & \multicolumn{4}{c}{NFCL-D}                                                                                \\ \cmidrule(lr){1-2} \cmidrule(lr){3-3} \cmidrule(lr){4-12} \cmidrule(lr){13-16}
Datasets                   & $T$        & -                                     & 16                                    & 32                                    & 64                                    & 128                                   & 16,16                                 & 32,32                                 & 64,64                                 & 128,128                               & 128,128,64                             & 64                                    & 128         & 64,64                                 & 128,128     \\ \cmidrule(lr){1-2} \cmidrule(lr){3-3} \cmidrule(lr){4-12} \cmidrule(lr){13-16} \morecmidrules \cmidrule(lr){1-2} \cmidrule(lr){3-3} \cmidrule(lr){4-12} \cmidrule(lr){13-16}
                           & 6          & 0.821                                 & 0.821                                 & 0.822                                 & 0.824                                 & {\color[HTML]{0000FF} \textbf{0.827}} & 0.822                                 & 0.824                                 & 0.823                                 & \ul{0.826}                           & 0.825                                  & 0.824                                 & 0.824       & 0.822                                 & 0.822       \\
\bottomrule \bottomrule
\end{tabular}%
}
\caption{$R^2$ ($\uparrow$) comparison among hyperparameter tuning.}
\label{tab:hp_search_r2}
\end{table*}

\section{Full output layer weight map}\label{full_weight_map}

\begin{figure*}[tp]
  \centering
  \includegraphics[width=.9\textwidth]{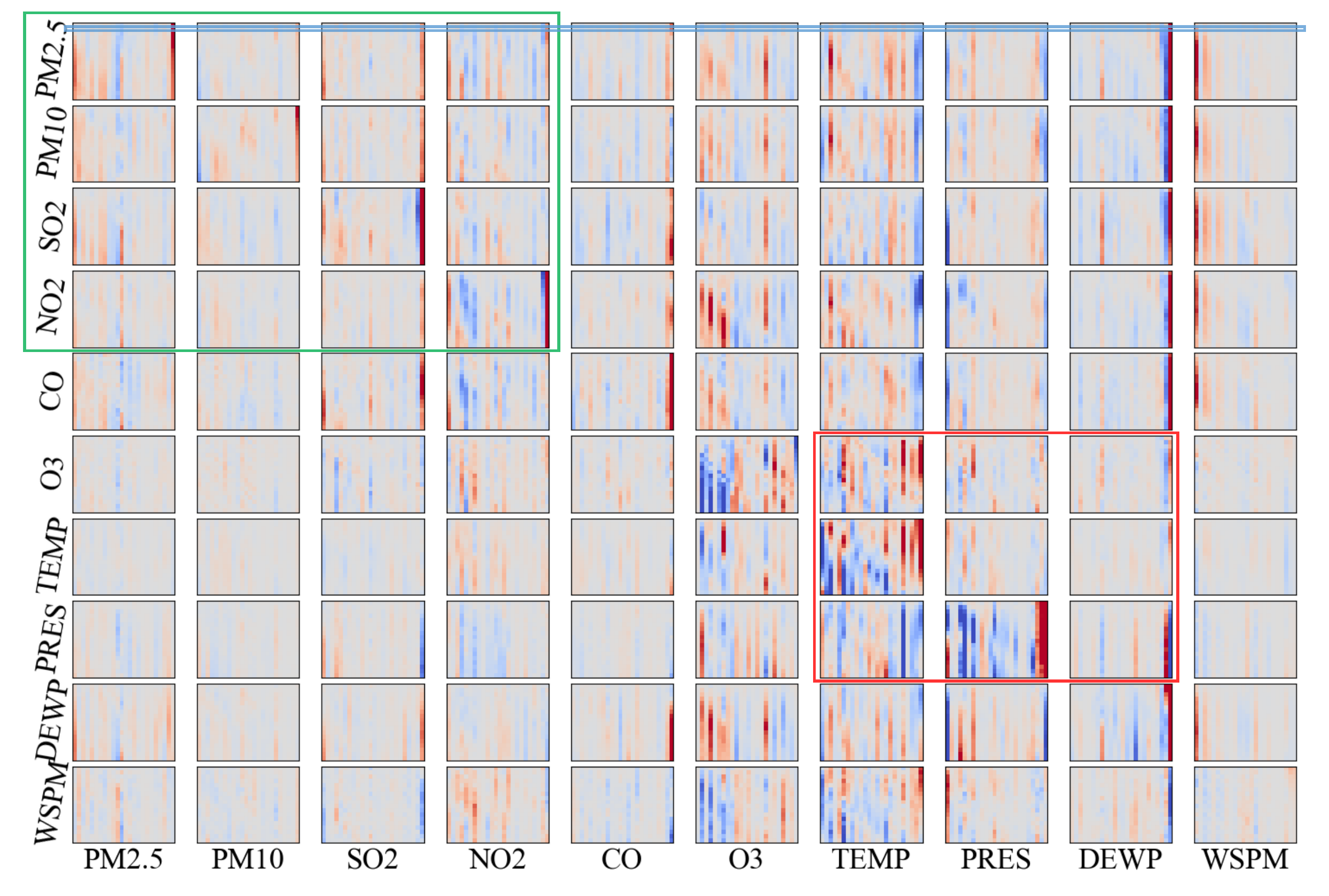}
  \caption{Fully illustrated the output layer's weight $h(X) \odot w$.}
  \label{fig.w-cdot-hx}
  \Description{  }
\end{figure*}

Every NFCL performs the operation $X' \cdot w$ (as defined in Section \ref{NFCL}, where $w \in \mathbb{R}^{k,l,k,t}$, and $X' \subset {X, h(X)}$).
The full map is obtained by reshaping $w$ into $\mathbb{R}^{k, l, kt}$, and then conducting an element-wise product with $X'$.
Finally, we extract the result of $k \cdot L + t$ rows from the full map to predict a specific time-series $k$ at a specific time $t$.
In the main context, the sub-sections \ref{sec:interpret} and \ref{sec:analysis_cont} used the figure \ref{fig.w-cdot-hx}.
The sub-sections \ref{sec:interpret} used the part of the blue-colored wide-shallow box in Figure \ref{fig.w-cdot-hx}. 
Moreover, blued boxes reshaped into $\mathbb{R}^{k, l}$ only to provide forecasting reasons in a specific time-series $k$ at a specific time $t$.
Also, the sub-sections \ref{sec:analysis_cont} used the part of the red and green colored boxes in Figure \ref{fig.w-cdot-hx}.

\section{Analysis of the number of model's parameters}\label{app:count_params}

Table \ref{tab:param_count} shows the number of learnable parameters.
From the first to the third columns denote the dataset, the number of time-series $K$, and the forecasting length $T$.
The fourth column onward shows the number of training parameters per data shape in the model specified in the first row.

\begin{table*}[t]
\centering
\resizebox{\textwidth}{!}{%
\begin{tabular}{cccccccccccccc} \toprule \toprule
Dataset & $K$ & $T$ & NFCL-V & NFCL-C & DLinear & NLinear & TiDE & Informer & Autoformer & PatchTST & iTransformer & SCINet & TimesNet \\ \cmidrule(lr){1-3} \cmidrule(lr){4-14}
\multirow{4}{*}{Weather} & \multirow{4}{*}{21} & 6 & 63672 & 112560 & 300 & 150 & 1585331 & 11375125 & 10604565 & 6324230 & 6321670 & 100104 & 1174579 \\
 &  & 12 & 127302 & 176190 & 600 & 300 & 1727297 & 11375125 & 10604565 & 6333452 & 6324748 & 100248 & 1174729 \\
 &  & 18 & 190932 & 239820 & 900 & 450 & 1869263 & 11375125 & 10604565 & 6342674 & 6327826 & 100392 & 1174879 \\
 &  & 24 & 254562 & 303450 & 1200 & 600 & 2011229 & 11375125 & 10604565 & 6351896 & 6330904 & 100536 & 1175029 \\\cmidrule(lr){1-3} \cmidrule(lr){4-14}
\multirow{4}{*}{Nong.} & \multirow{4}{*}{10} & 6 & 14480 & 37760 & 300 & 150 & 1496486 & 11338762 & 10551306 & 6324230 & 6321670 & 23104 & 1173912 \\
 &  & 12 & 28940 & 52220 & 600 & 300 & 1570604 & 11338762 & 10551306 & 6333452 & 6324748 & 23248 & 1174062 \\
 &  & 18 & 43400 & 66680 & 900 & 450 & 1644722 & 11338762 & 10551306 & 6342674 & 6327826 & 23392 & 1174212 \\
 &  & 24 & 57860 & 81140 & 1200 & 600 & 1718840 & 11338762 & 10551306 & 6351896 & 6330904 & 23536 & 1174362 \\\cmidrule(lr){1-3} \cmidrule(lr){4-14}
\multirow{4}{*}{Exchange} & \multirow{4}{*}{8} & 6 & 9280 & 27904 & 300 & 150 & 1479538 & 11330568 & 10540040 & 6324230 & 6321670 & 14928 & 1173766 \\
 &  & 12 & 18544 & 37168 & 600 & 300 & 1541320 & 11330568 & 10540040 & 6333452 & 6324748 & 15072 & 1173916 \\
 &  & 18 & 27808 & 46432 & 900 & 450 & 1603102 & 11330568 & 10540040 & 6342674 & 6327826 & 15216 & 1174066 \\
 &  & 24 & 37072 & 55696 & 1200 & 600 & 1664884 & 11330568 & 10540040 & 6351896 & 6330904 & 15360 & 1174216 \\\cmidrule(lr){1-3} \cmidrule(lr){4-14}
\multirow{4}{*}{ILL} & \multirow{4}{*}{7} & 6 & 7112 & 23408 & 300 & 150 & 1471321 & 11326983 & 10534919 & 6324230 & 6321670 & 11512 & 1173701 \\
 &  & 12 & 14210 & 30506 & 600 & 300 & 1526935 & 11326983 & 10534919 & 6333452 & 6324748 & 11656 & 1173851 \\
 &  & 18 & 21308 & 37604 & 900 & 450 & 1582549 & 11326983 & 10534919 & 6342674 & 6327826 & 11800 & 1174001 \\
 &  & 24 & 28406 & 44702 & 1200 & 600 & 1638163 & 11326983 & 10534919 & 6351896 & 6330904 & 11944 & 1174151 \\ \bottomrule \bottomrule
\end{tabular}%
}
\caption{The number of learnable parameters according to the nine baselines and two NFCLs. Note that NFCL-D has the number of learnable parameter as three times of NFCL-C's since we hired the three decomposition blocks.}
\label{tab:param_count}
\end{table*}

\section{Comparison of NFCLs generalization.}\label{app:nfcl_generalization}

\begin{figure*}[tp]
  \centering
  \includegraphics[width=.8\textwidth]{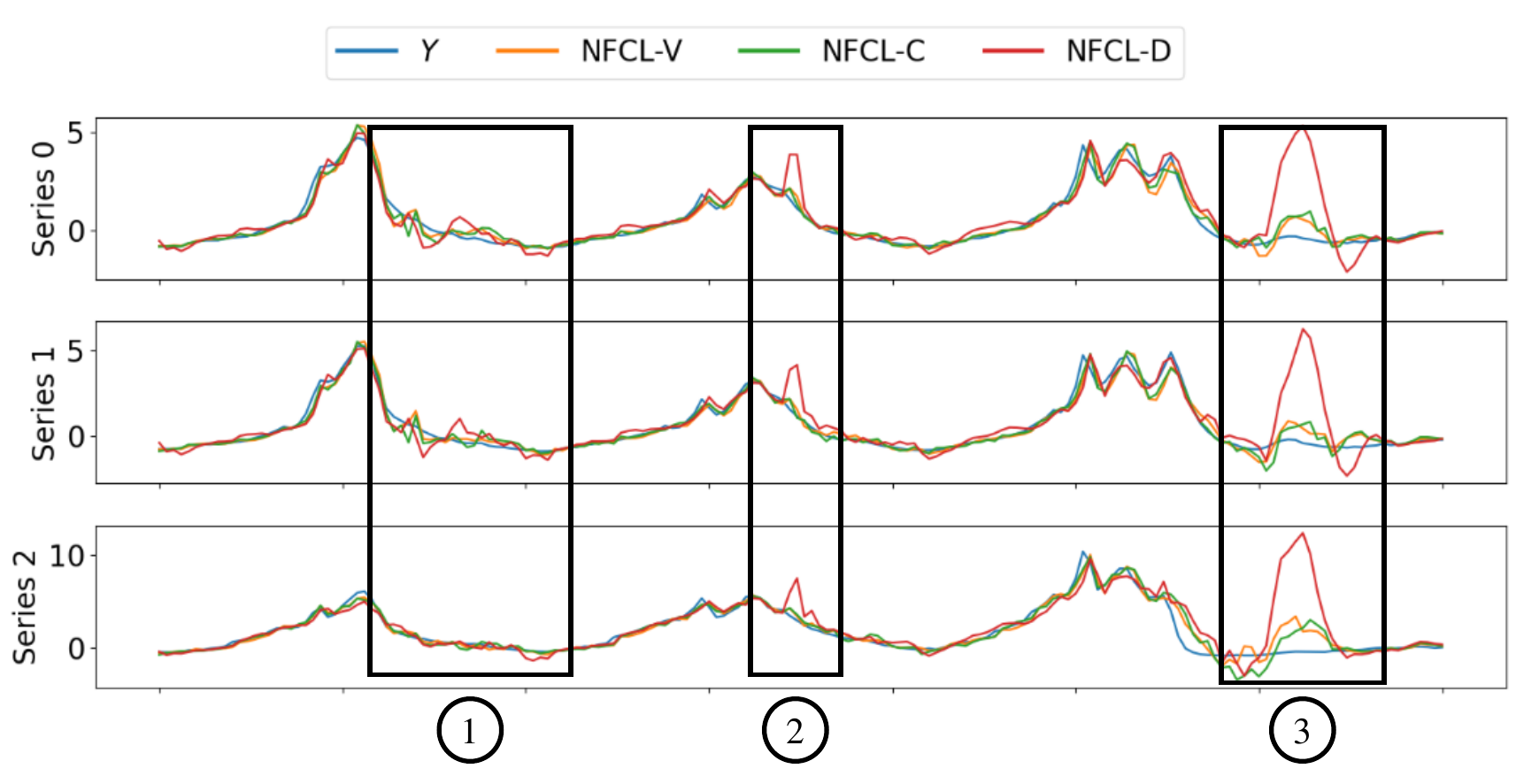}
  \caption{Illustration of NFCLs forecasting when $T=0$.}
  \label{fig.nfcl_general_test}
  \Description{  }
\end{figure*}

We meticulously prepared figures and tables to systematically analyze the performance degradation of NFCL-D.
Initially, Figure \ref{fig.nfcl_general_test} was constructed by training NFCL-V, C, and D adequately on the ILL dataset, and NFCLs set up the $T$ to 18. 
Subsequently, upon completion of the training process, only the values corresponding to $T=0$ among the predicted results for the test set were imported and visualized. 
In the figure \ref{fig.nfcl_general_test}, the blue, yellow, green, and red lines represent the real $Y$ and the predictions of NFCL-V, C, and D, respectively.

Figure \ref{fig.nfcl_general_test} illustrates three instances of prediction failure. Firstly, in area 1, NFCL-D exhibits a failure in prediction compared to the successful predictions of NFCL-V and C. However, NFCL-D demonstrates similar prediction results to other models for series 2. In regions 2 and 3, NFCL-D demonstrates incorrect pattern predictions. Furthermore, in Region 3, all NFCL variants display inaccurate time-series forecast results, with NFCL-D notably exhibiting a larger error compared to the other NFCLs.

\begin{table*}[t]
\centering
\begin{tabular}{ccccccc} \toprule \toprule
Dataset & Split & Variation & MAE & MSE & SMAPE & $R^2$ \\ \cmidrule(lr){1-3} \cmidrule(lr){4-7}
\multirow{9}{*}{ILL} & \multirow{3}{*}{Train} & NFCL-V & 0.390 & 0.387 & 37.846 & 0.629 \\
 &  & NFCL-C & 0.288 & 0.224 & 30.394 & 0.785 \\
 &  & NFCL-D & 0.287 & 0.213 & 30.451 & 0.796 \\\cmidrule(lr){2-3} \cmidrule(lr){4-7}
 & \multirow{3}{*}{Val} & NFCL-V & 0.453 & 0.472 & 40.218 & 0.366 \\
 &  & NFCL-C & 0.366 & 0.288 & 32.873 & 0.599 \\
 &  & NFCL-D & 0.378 & 0.290 & 33.719 & 0.582 \\\cmidrule(lr){2-3} \cmidrule(lr){4-7}
 & \multirow{3}{*}{Test} & NFCL-V & 1.261 & 5.197 & 42.969 & -0.211 \\
 &  & NFCL-C & 1.320 & 7.211 & 40.082 & -0.680 \\
 &  & NFCL-D & 1.497 & 11.411 & 40.067 & -1.569 \\ \cmidrule(lr){1-3} \cmidrule(lr){4-7}
\multirow{9}{*}{Weather} & \multirow{3}{*}{Train} & NFCL-V & 0.171 & 0.276 & 20.017 & 0.724 \\
 &  & NFCL-C & 0.159 & 0.250 & 19.656 & 0.750 \\
 &  & NFCL-D & 0.163 & 0.257 & 19.924 & 0.743 \\\cmidrule(lr){2-3} \cmidrule(lr){4-7}
 & \multirow{3}{*}{Val} & NFCL-V & 0.189 & 0.348 & 20.431 & 0.706 \\
 &  & NFCL-C & 0.183 & 0.319 & 20.491 & 0.730 \\
 &  & NFCL-D & 0.186 & 0.328 & 20.741 & 0.723 \\\cmidrule(lr){2-3} \cmidrule(lr){4-7}
 & \multirow{3}{*}{Test} & NFCL-V & 0.109 & 0.091 & 16.712 & 0.704 \\
 &  & NFCL-C & 0.108 & 0.089 & 16.691 & 0.716 \\
 &  & NFCL-D & 0.110 & 0.090 & 16.975 & 0.711 \\ \bottomrule \bottomrule
\end{tabular}%
\caption{The train-validation-test performance comparison according to NFCLs.}
\label{tab:generalization_compare}
\end{table*}

A glimpse into the cause of NFCL-D's poor performance can also be obtained from Table \ref{tab:generalization_compare}. The first row of the table outlines the dataset used for learning, the data division method, and the type of NFCL employed. NFCL utilized one layer and 32 hidden units, while NFCL-D's decomposition was configured as {10,4,1}. Furthermore, each experiment was conducted only once.

Results from the experiments indicate that in the case of the Weather dataset, which contains a substantial amount of data, both NFCL-C and NFCL-D consistently exhibit performance enhancements over NFCL-V across all training, validation, and evaluation datasets. However, when considering the ILL dataset, NFCL-D demonstrated notably strong performance during training but displayed poor performance indicators during testing, specifically on the evaluation dataset. This discrepancy highlights an instance of NFCL-D overfitting to the training data, as evidenced by inappropriate pattern predictions (as illustrated in Figure \ref{fig.nfcl_general_test}). 
Consequently, it becomes evident that NFCLs can effectively capture the data distribution given a sufficient volume of data.

\end{document}